\documentclass[final]{cvpr}
\usepackage{times}
\usepackage{epsfig}
\usepackage{graphicx}
\usepackage{amsmath}
\usepackage{amssymb}
\usepackage{rotating}
\usepackage{csquotes}
\usepackage{multirow}
\usepackage{bbding}
\usepackage[table,xcdraw]{xcolor}
\usepackage[normalem]{ulem}

\usepackage[pagebackref=true,breaklinks=true,colorlinks,bookmarks=false]{hyperref}

\def\blu#1{\textbf{\color{blue} #1}} 
\def\red#1{\textbf{\color{red}  #1}} 

\def\cvprPaperID{766} 
\def\confYear{CVPR 2021}

\begin{document}

\title{Weakly Supervised Video Salient Object Detection}
\author{
Wangbo Zhao$^{1}$\quad
Jing Zhang$^{2,3}$\quad
Long Li$^{^1}$\quad
Nick Barnes$^{2}$\quad
Nian Liu$^{4}$\quad
Junwei Han$^1${\Envelope}~\thanks{Corresponding author: Junwei Han \emph{(junweihan2010@gmail.com)}}\quad\\
$^1$ The Brain and Artificial Intelligence Laboratory, Northwestern Polytechnical University \quad \\
$^2$ Australian National University \quad
$^3$ CSIRO, Australia\\
$^4$ Inception Institute of Artificial Intelligence \\
\{wangbo.zhao96, zjnwpu, longli.nwpu,  liunian228, junweihan2010\}@gmail.com, \\
nick.barnes@anu.edu.au
}








\maketitle



\begin{abstract}
Significant performance improvement has been achieved for fully-supervised video salient object detection with the pixel-wise labeled training datasets, which are
time-consuming and expensive to obtain.
To relieve the burden of data annotation, we present the first weakly supervised video salient object detection model based on relabeled
\enquote{fixation guided scribble annotations}.
Specifically, an \enquote{Appearance-motion fusion module} and bidirectional ConvLSTM based framework are proposed to achieve effective multi-modal learning and long-term temporal context modeling based on our new weak annotations.
Further, we design a novel foreground-background similarity loss to further explore the labeling similarity across frames. A weak annotation boosting strategy is also introduced to boost our model performance with a new pseudo-label generation technique.
Extensive experimental results on six benchmark video saliency detection datasets illustrate the effectiveness of our solution\footnote{Our code and data is publicly available at: \url{https://github.com/wangbo-zhao/WSVSOD}.}.

\end{abstract}

\vspace{-4mm}
\section{Introduction}
Video salient object detection (VSOD) models are designed to segment salient objects in both the spatial domain and the temporal domain. Existing VSOD methods focus on two different solutions: 1) encoding temporal information using a recurrent network \cite{song2018pyramid,fan2019shifting,yan2019semi}, \eg LSTM; and 2) encoding geometric information using the optical flow constraint \cite{li2019motion, ren2020tenet}. Although considerable performance improvements have been achieved,
we argue that the huge burden of pixel-wise labeling makes VSOD a much more expensive task than the RGB image-based saliency detection task \cite{hou2017deeply,liu2020picanet,liu2019simple, zhao2019egnet, wu2019stacked}.


\begin{figure}[!htp]
   \begin{center}
   \begin{tabular}{{c@{ } c@{ } c@{ }}}
   {\includegraphics[width=0.28\linewidth]{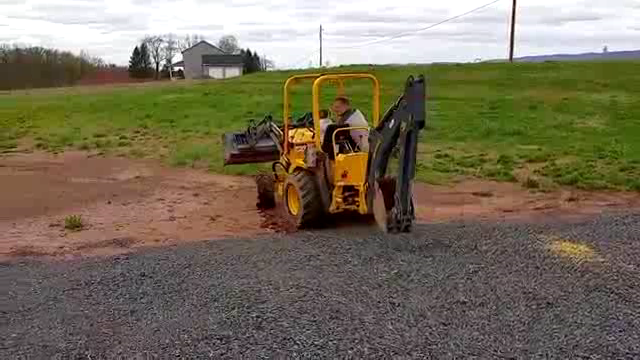}}&
    {\includegraphics[width=0.28\linewidth]{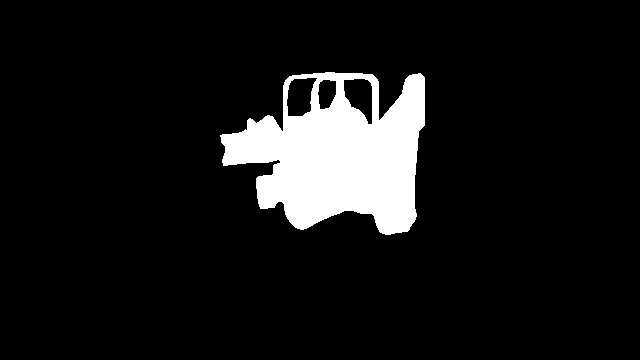}}&
    {\includegraphics[width=0.28\linewidth]{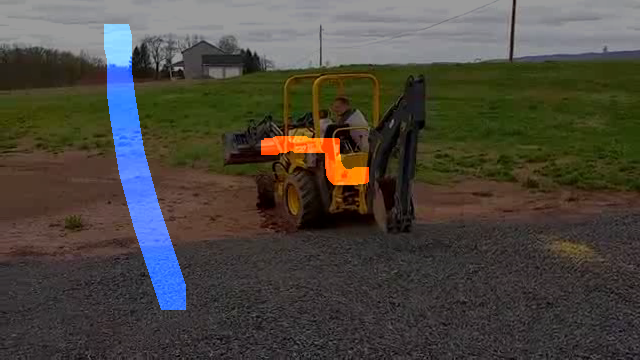}} \\
    \footnotesize{(a) Image} &
    \footnotesize{(b) Full Anno.} &
    \footnotesize{(c) Weak Anno.} \\
     {\includegraphics[width=0.28\linewidth]{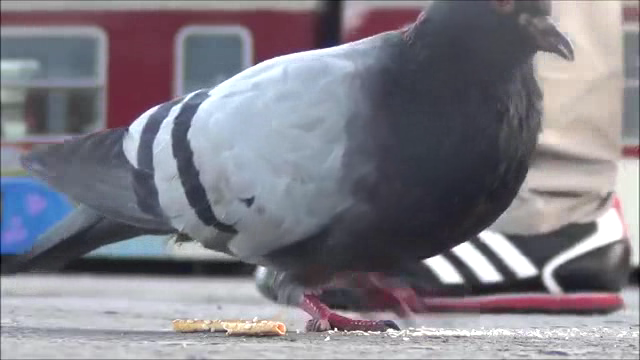}}&
    {\includegraphics[width=0.28\linewidth]{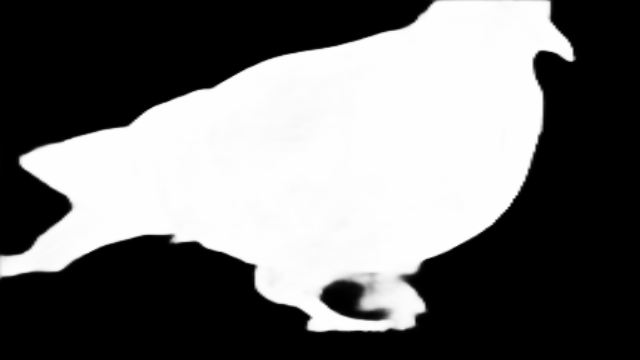}}&
    {\includegraphics[width=0.28\linewidth]{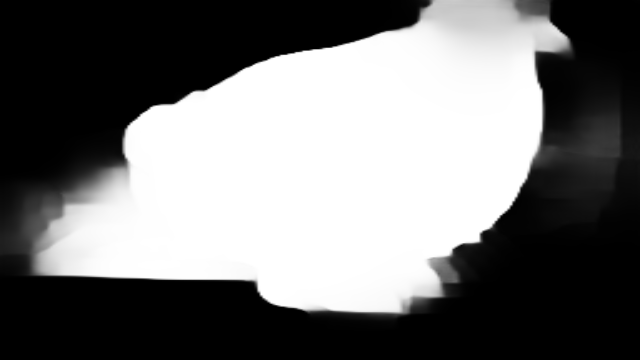}}\\
    \footnotesize{(d) Image} &
    \footnotesize{(e) TENet\cite{ren2020tenet}}  & \footnotesize{(f) Ours}\\
   \end{tabular}
   \end{center}
   \caption{Training with our
   weak annotation (c), we achieve competitive performance (f) compared with
   TENet\cite{ren2020tenet} (e).}
\label{fig:performance_compare_figure1}
\end{figure}

The standard pipeline to train a deep video saliency detection model involves two main steps. Firstly, the network is pre-trained on an existing static RGB image-based saliency detection training dataset, \eg DUTS \cite{duts_train_set} or MSRA10K \cite{msra10k_train_set}. Then, it is fine-tuned on video saliency detection datasets, \eg DAVSOD \cite{fan2019shifting} and DAVIS \cite{perazzi2016benchmark}.
The main reason for using this strategy is that video saliency datasets usually have limited scene diversity. Although the largest DAVSOD dataset \cite{fan2019shifting} has more than 10K frames for training, the large redundancy
across
the frames of each clip makes it still insufficient to effectively train deep video saliency models.
Specifically,
DAVSOD has a total of 107 clips for training and validation, which only indicates around 107 diverse scenes. Hence, directly training with a VSOD dataset may lead to poor model generalization ability, as the model may overfit on the highly redundant data.

To obtain an effective video saliency detection model, existing fully supervised VSOD methods \cite{li2019motion, ren2020tenet,fan2019shifting} rely on both RGB image saliency datasets and VSOD training datasets.
The problem behind the above pipeline is the huge requirement for pixel-wise labeling, which is time-consuming and expensive to obtain. For example, RGB image saliency training datasets have more than 10K labeled samples \cite{duts_train_set,msra10k_train_set}. Further, as shown in Tab.~\ref{tab:dataset}, widely used VSOD training datasets (DAVSOD and DAVIS) contain
more than 14K pixel-wise labeled frames. Both of them required large burden to perform data annotations.

\begin{table}[t!]
  \centering
  \scriptsize
  \renewcommand{\arraystretch}{1.1}
  \renewcommand{\tabcolsep}{1.4mm}
  \caption{\small Details of 
  existing video sod datasets. Dataset: name of the dataset, Size: number of frames, Annotated size: 
  labeled frames(per pixel), Training: Frames used for training, /: this dataset is not split.
  }\label{tab:dataset}
  \begin{tabular}{l|c|c|c|c}
  \hline
    \textbf{Dataset}    & Released Year
                       & Size
                       & Annotated size
                       & Training
      
                       \\

  \hline
    DAVSOD\cite{fan2019shifting}  &2019  &23,938   &23,938   & 12,670\\
  \hline
  
  \hline
    VOS\cite{li2017benchmark}     &2018  &116,103  &7,467 & 5,927 \\
  \hline
  
  \hline
    DAVIS\cite{perazzi2016benchmark}  &2016  &3,455  &3,455 & 2,079 \\
  \hline
  
  \hline
    ViSal\cite{wang2015consistent}   &2015  &963    &193 & /\\
  \hline
  
  \hline
    FBMS\cite{ochs2013segmentation}   &2014  &13,860  &720 & 353\\
    
  \hline
   SegV2\cite{li2013video}            &2013   &1,065   &1,065 & /\\
  \hline

  \end{tabular}
  \vspace{-4mm}
\end{table}

To relieve the burden of pixel-wise labeling, one can resort the
weakly supervised learning technique \cite{zhang2020weakly,duts_train_set} to learn saliency from image scribble or image-level labels. In this paper, considering the efficiency of scribble annotation, we aim to learn a weakly supervised video saliency detection network via scribble. However, the main problem is that the per-image labeled scribble has no temporal information. To incorporate temporal information into our weak annotation, we adopt the fixation annotation in existing VSOD training datasets
as guidance, and propose fixation guided scribble annotation as shown in Fig.~\ref{fig:performance_compare_figure1} (c). Specifically, we first define the regions that have the peak response of fixation as foreground and those without fixation as background. Then we label both foreground scribble and background scribble following \cite{zhang2020weakly}.




Based on the fixation guided scribble annotation, we design an appearance-motion fusion module to fuse both appearance information from the RGB image and motion information from 
optical flow as shown in Fig. \ref{network_overview}. Furthermore, a bidirectional LSTM \cite{song2018pyramid} based temporal information enhanced module is presented to further obtain long-term temporal information. Note that, we use scribble annotation from S-DUTS \cite{duts_train_set} to pre-train our video saliency detection network as the conventional way. Build upon both scribble annotation from the RGB image saliency dataset and video saliency dataset, our weakly supervised video saliency detection network leads to a very cheap configuration compared with existing deep video saliency detection models.
Moreover, considering the cross-frame redundancy of the video saliency dataset, we introduce the foreground-background similarity loss to fully explore our weak annotation. We also introduce a weak annotation boosting strategy by leveraging our scribble annotation and the saliency map generated from the off-the-shelf fully-supervised SOD model. Benefiting from these, our model can achieve comparable results with state-of-the-art fully-supervised methods. \eg Fig.~\ref{fig:performance_compare_figure1} (f) and (e).



Our main contributions are: 1) We introduce the first weakly supervised video salient object detection network based on our fixation guided scribble annotation; 2) We propose an appearance-motion fusion module and a temporal information enhance module to effectively fuse appearance and motion features; 3) We present the foreground-background similarity loss to explore our weak annotation in adjacent frames; 4) We combine saliency maps generated from an off-the-shelf saliency model and our scribble annotations to further boost model performance.




  


\section{Related Work}

\noindent\textbf{Fully supervised video salient object detection:}
As the mainstream of video salient object detection, the fully supervised video saliency detection models mainly focus on
exploring both spatial and temporal information of the training dataset.
Wang \etal \cite{wang2017video} models the short-term spatial-temporal information by taking two adjacent frames as input. To model the longer spatio-temporal information, \cite{song2018pyramid, li2018flow} adopt ConvLSTM to capture richer spatial and temporal features simultaneously. Some methods also model the human attention mechanism to select interesting regions in different frames, \eg self-attention \cite{gu2020pyramid}, spatial attention supervised by human eye ﬁxation data \cite{fan2019shifting, wang2020paying}. As objects with motion in a video are usually salient, li \etal \cite{li2019motion} use optical flow as guidance to find the salient regions. Ren \etal \cite{ren2020tenet} combine the spatial and temporal information and present a semi-curriculum learning strategy to reduce the learning ambiguities.

While these methods show their successes on VSOD, they heavily rely on the large densely annotated training datasets. Annotating a high-quality dataset is expensive and time-consuming.
Different from them, depending on only weak annotations, our method greatly relief the labeling burden, which is both cheaper and more accessible. 


\noindent\textbf{Weakly/semi/un-supervised video salient object detection:}
There are many traditional unsupervised VSOD methods \eg \cite{wang2015saliency, wang2015consistent, li2017benchmark}, most of which exploit handcrafted features, which
makes them unsatisfactory in the real-world application. When it comes to the learning-based methods, although the problem of depending on laborious and costly pixel-wise dense annotation is obvious and serious, few methods make an effort to alleviate it. To the best of our knowledge, there is no previous method to solve VSOD with totally weakly labeled data. There are only several methods that try to use less annotated data. Yan \etal \cite{yan2019semi}
addresses VSOD in a semi-supervised manner by using pseudo labels, where the optical flow map serves as guidance for pseudo label generation with the
sparsely annotated frames. Tang \etal \cite{tang2018weakly} uses limited manually labeled data and pseudo labels generated from existing saliency models to train their model. Recently weakly-supervised finetuning during testing is also explored. Li \etal \cite{li2020plug} proposes to generate pseudo labels to weakly retrain pre-trained saliency models during testing. But high-quality labeled data is still inevitable to obtain the pseudo labels. 

Different from previous methods, which rely on all or part of the fully annotated dataset, our model is end-to-end trainable without using
any densely annotated labels.

\noindent\textbf{Video object segmentation:}
There are two types of video object segmentation(VOS) models, including Zero-shot VOS
\cite{wang2020paying, song2018pyramid, zhou2020motion} and One-shot VOS
\cite{grundmann2010efficient, caelles2017one, wang2019ranet}. During testing, the former aims at segmenting primary objects in a video without any help, while the first annotated frame is given in the later. Since Zero-shot VOS is more similar to VSOD, we only discuss the related literature in this paper. Among them, Song \etal \cite{song2018pyramid} solve VOS and VSOD at the same time. Wang \etal \cite{wang2019zero} builds a fully connected graph to mine rich and high-order relations between video frames. Zhou \etal \cite{zhou2020motion} proposes a motion attention block to leverage motion information to reinforce spatio-temporal object representation. \cite{mahadevan2020making} introduces an encoder-decoder network consisting entirely of 3D convolutions.

Like VSOD, most of the video segmentation models are fully supervised, and
the weakly-supervised or semi-supervised counterparts is still under-explored. Until recently, Lu \etal \cite{lu2020learning} introduces the intrinsic properties of VOS at multiple granularities to learn from weak supervision. However, the undesirable performance and slow inference speed makes it desirable to further explore this task.
\section{Our Method}
\subsection{Overview}

\begin{figure*}[!t]
  \graphicspath{{figure1/network/}}
  \centering
  \includegraphics[width=0.80\linewidth]{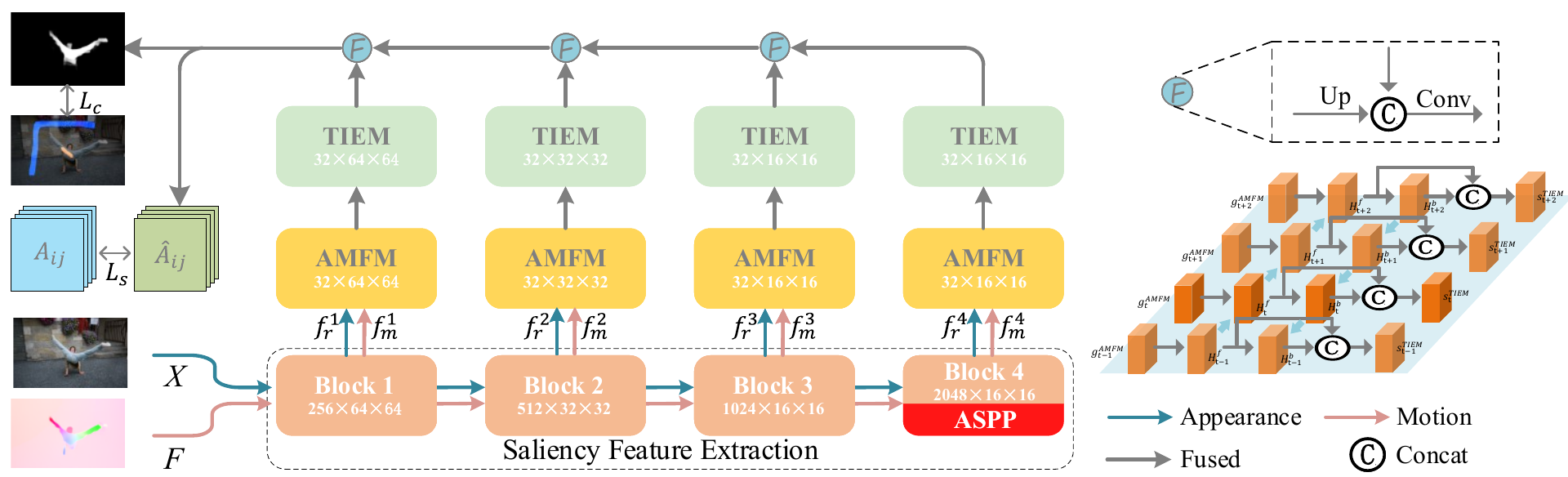}
  \caption{Overview of the proposed model. For simplicity, we do not show the edge detection branch borrowed from \cite{zhang2020weakly} here. Details about TIEM can be found on the right. There is no upsample operation in the first "F"."Up":"Upsample operation; "C": concatenation operation;"Conv":$3\times3$ convolutional layer.}
  \label{network_overview}
\end{figure*}

\begin{figure}[!htp]
   \begin{center}
   \begin{tabular}{{c@{ } c@{ } c@{ } c@{ }}}
   {\includegraphics[width=0.22\linewidth]{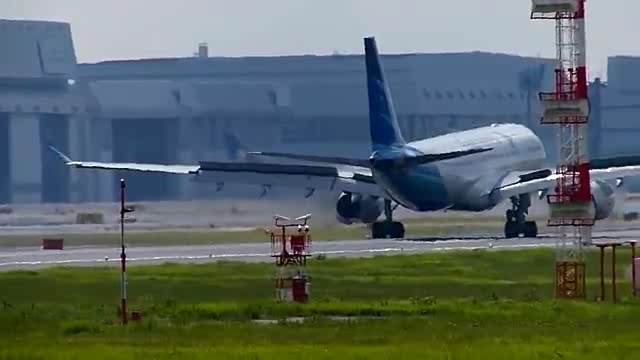}}&
    {\includegraphics[width=0.22\linewidth]{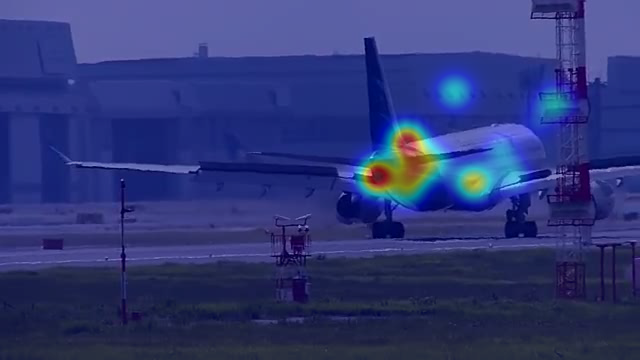}}&
    {\includegraphics[width=0.22\linewidth]{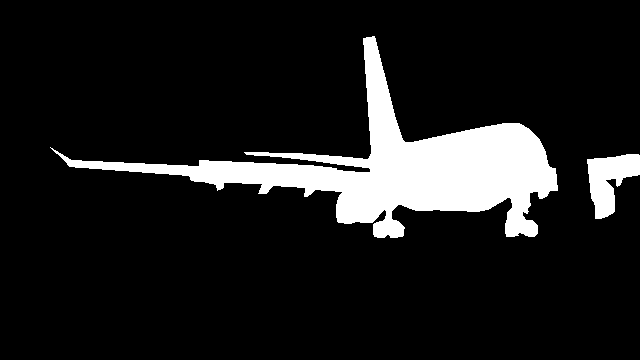}}&
    {\includegraphics[width=0.22\linewidth]{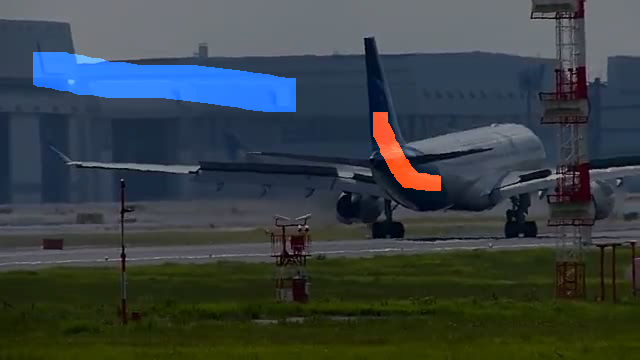}} \\
    {\includegraphics[width=0.22\linewidth]{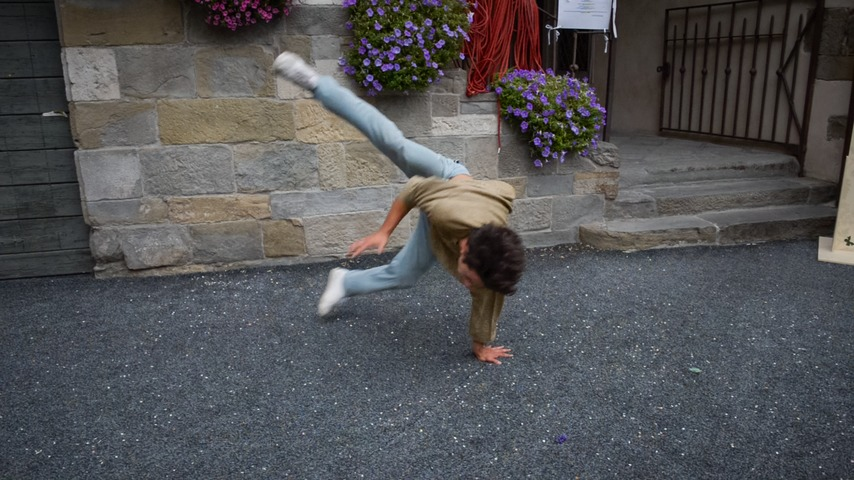}}&
    {\includegraphics[width=0.22\linewidth]{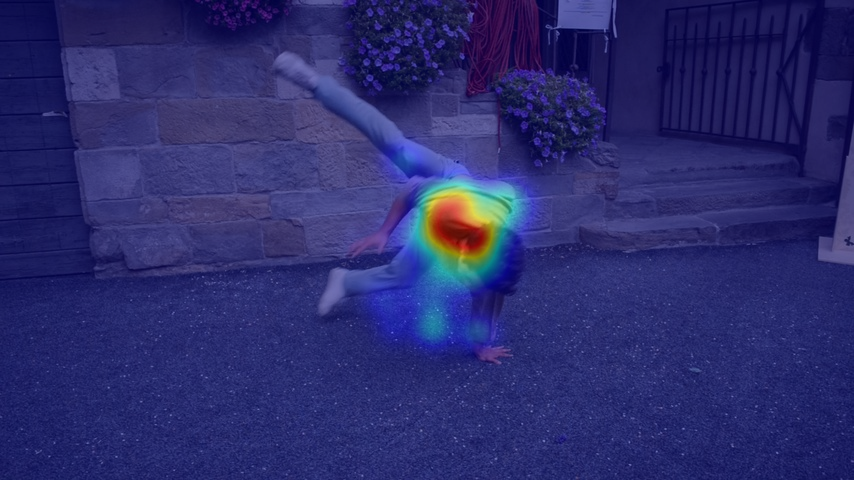}}&
    {\includegraphics[width=0.22\linewidth]{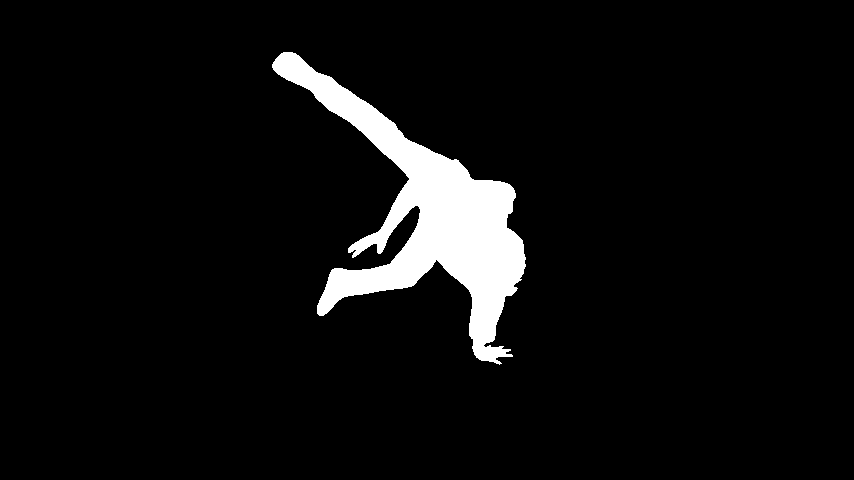}}&
    {\includegraphics[width=0.22\linewidth]{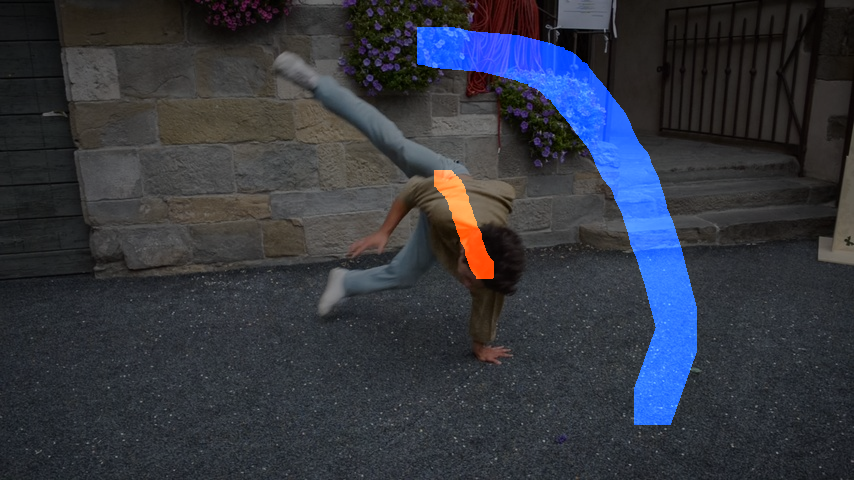}} \\
    \footnotesize{(a) Image} &
    \footnotesize{(b) Fixation} & \footnotesize{(c) Clean GT} & \footnotesize{(d) Our GT} \\
   \end{tabular}
   \end{center}
   \caption{The fixation guided scribble annotation, where we obtain the sequence of scribble labels(d) with the temporal information from fixation annotation(b).
   }
\label{fig:fixation_guided_scribble}
\end{figure}

As a weakly supervised video saliency detection framework, we first relabel existing video saliency detection datasets DAVSOD \cite{fan2019shifting} and DAVIS \cite{perazzi2016benchmark} with scribble labels. Due to a lack of temporal information in the per-image scribble annotation, we introduce fixation guided scribble annotation as shown in Fig.~\ref{fig:fixation_guided_scribble}.  Our training dataset is then defined as $T=\{X, F, Y\}$, where $X$ is the RGB image, $F$ is the optical flow map predicted from \cite{sun2018pwc}, $Y$ is our fixation guided scribble annotation.
We first design a \emph{Saliency feature extraction} $f_\alpha$ to extract features $f_\alpha(X)$ and $f_\alpha(F)$ from RGB images and flow respectively, where $\alpha$ is the network parameter set.
Then, we present the \emph{Appearance-Motion Fusion Module} (AMFM) $g_\beta(f_\alpha(X),f_\alpha(F))$ to effectively learn from both appearance information (\eg the RGB image branch) and motion information (\eg the flow branch). Further, we introduce a \emph{Temporal Information Enhanced Module} (TIEM) $s_\gamma(g_\beta)$ by using ConvLSTM to model the long-term temporal information between frames.  For each frame, we  fuse  features from  TIEM  in  different  levels  in  a top-down  manner  to  get the final output. To further explore the temporal information from our fixation guided scribble annotation, we present a \emph{Foreground-background similarity loss} as a frame-wise constraint.
Moreover, we present a \emph{Saliency boosting strategy} to further improve the performance of our method. An overview of our network is shown in Fig.~\ref{network_overview}.

\subsection{Fixation guided scribble annotation}

The largest video saliency detection dataset, \ie DAVSOD \cite{fan2019shifting}, is annotated in two steps: 1) an eye tracker is used to record fixation points, the output of which is then Gaussian blurred to obtain a dense fixation map; and 2) annotators segment the whole scope of the salient foreground based on the peak response region\footnote{The peak response region is the region with the densest fixation points.}. As indicated in \cite{fan2019shifting}, the extra fixation annotation introduces useful temporal information to the video saliency dataset.  Conventionally, DAVSOD is combined with the DAVIS \cite{perazzi2016benchmark} dataset to train fully supervised VSOD models.
Originally, DAVIS had no fixation annotation, however, it was added by Wang \etal \cite{wang2020paying}. As a weak video saliency detection network, we intend to use the fixation data as guidance to obtain temporal information, and we then replace the pixel-wise clean annotation with scribble for weakly supervised learning. Given every frame in our video saliency training dataset, as shown in Fig.~\ref{fig:fixation_guided_scribble} (a), and the corresponding fixation map as shown in Fig.~\ref{fig:fixation_guided_scribble} (b), we annotate the foreground scribble in the objects with peak response regions, and background scribble in other region as shown in Fig.~\ref{fig:fixation_guided_scribble} (d). In this case, the generated scribble annotation encodes temporal information, which is different from \cite{zhang2020weakly} where the scribble is totally image-based, with no temporal information.

\subsection{Saliency feature extraction}
As shown in Fig.~\ref{network_overview}, the saliency feature extraction module
is used to extract the appearance saliency feature $f_\alpha(X)$ from the RGB image $X$ and motion saliency feature $f_\alpha(F)$ from the optical flow map $F$. We build our architecture upon ResNet-50 \cite{he2016deep} and remove the down-sampling operations in stage four\footnote{We define a group of convolutional layers of the same spatial size as belonging to the same stage.} to keep the spatial information. Apart from this, we replace the convolutional layers in the last layer with dilated convolutions \cite{yu2015multi} with a dilated rate of 2. An ASPP \cite{chen2017rethinking} module is added after stage four to extract multi-scale spatial features, which includes one $1\times1$ convolutional layer, and three $3\times3$ dilated convolutional layers with dilation rate of 6, 12, and 18, and a global average pooling operation. With our saliency feature extraction module, we obtain the appearance feature  $f_\alpha(X)=\{f_r^1,f_r^2,f_r^3,f_r^4\}$
and motion feature $f_\alpha(F)=\{f_m^1,f_m^2,f_m^3,f_m^4\}$ respectively, where $f_r^k$ and $f_m^k$ are the appearance feature and motion feature of the $k$-th stage of the network, and $k$ indexes network stages. More details can be found in Section~\ref{experimental_results}. We also add an extra edge detection branch to recover the structure information of the final output, and details of which can be found in \cite{zhang2020weakly}.

\subsection{Appearance-motion fusion module}
\begin{figure}[!t]
  \graphicspath{{figure1/network/}}
  \centering
  \includegraphics[width=0.80\linewidth]{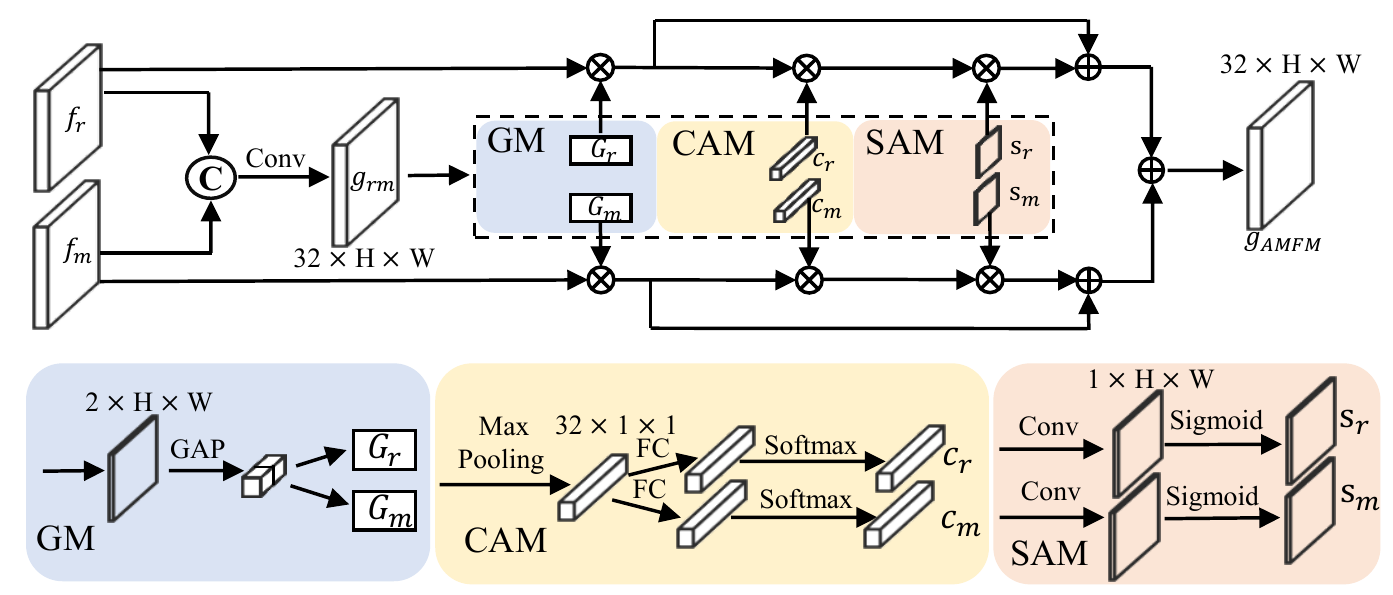}

  \caption{Appearance-motion fusion module. GM, CAM and SAM are the gate module, channel
  and spatial attention module respectively. Their outputs are $G$, $CA$ and $SA$, respectively.}
  \label{fusion}
\end{figure}

The appearance-motion fusion module aims to effectively fuse the appearance feature $f_\alpha(X)$ and motion feature $f_\alpha(F)$.
As shown in Fig. \ref{fusion}, the inputs of the AMFM are the appearance feature $f_r^k$ and the motion feature$f_m^k$ of size $C \times W \times H$.
We use two convolutional layers with a ReLU activation function to reduce the number of channels of $f_r^k$ and $f_m^k$ to $C=32$ respectively. Then the concatenation operation and a $1\times1$ convolutional layer is adopted to obtain the fused feature $g_{rm}^k$ of size $C \times W \times H$,
which contains the appearance and motion information. We use $g_{rm}$ instead of $g_{rm}^k$ in the following for simplicity. 

There exists three sub-modules in our AMFM, namely the gate module (GM), the channel attention module (CAM) and the spatial attention module (SAM). The gate module is designed to control the importance of appearance features and motion features, and the two attention modules are used to select the discriminative channels and locations. In GM, two different gates can be generated from $g_{rm}$, namely the appearance gate $G_r(g_{rm})$ and motion gate $G_m(g_{rm})$.
This module is designed to control the importance of $f_r$ and $f_m$, which is
formulated as:
\begin{equation} 
\boldsymbol{G} = GAP(\sigma(Conv(g_{rm};\beta))),
\end{equation}
where $\boldsymbol{G}=[G_r, G_m]$, and $G_{r}, G_{m}$ are two scalars in the range $[0,1]$. $Conv(g_{rm}; \beta)$ is a $1\times1$ convolutional layer, which reduces the channels of feature $g_{rm}$ from $C$ to $2$. $GAP(*)$ is the global average pooling layer in the spatial dimension, and $\beta$ is the network parameter set. $\sigma(*)$ is the sigmoid function.

The gate module produces two different scalars, representing the importance of appearance information and motion information. However, it can not emphasize important channels and spatial locations in appearance and motion features. Based on this, we propose our two attention modules, namely CAM and SAM, as:
\begin{equation} 
\boldsymbol{CA} = Softmax(FC(MaxPooling(g_{rm}); \beta)),
\end{equation}
\begin{equation} 
\boldsymbol{SA} = \sigma(Conv(g_{rm}; \beta)),
\end{equation}
where $\boldsymbol{CA}=[c_{r}, c_{m}]$
are the two
channel attention maps of size $C \times 1 \times 1$ for appearance and motion. $MaxPooling(*)$ is in the spatial dimensions. $FC$ is a fully connected layer with $2C$ output channels. The $Softmax$ function is implemented in every $C$ channels along the channel dimension. $\boldsymbol{SA}=[s_{r}, s_{m}]$, and $s_{r}, s_{m}$ are two
spatial attention maps of size $1 \times W \times H$.
Subsequently, the obtained gates $[G_{r}, G_{m}]$, channel attention tensors $[c_{r}, c_{m}]$, spatial attention tensors $[s_{r}, s_{m}]$ can be multiplied with $f_r$ and $f_m$
respectively to achieve both importance reweighting (the gate module) and attention reweighting (the attention modules). However, such a simple multiplication approach may lose some useful information. Inspired by \cite{li2019motion, wang2019learning}, we use the gated feature in a residual form as:
\begin{equation} 
g_{r} = (G_{r}\otimes f_{r})(1+s_{r}\otimes c_{r}),
\end{equation}
\begin{equation} 
g_{m} = (G_{m}\otimes f_{m})(1+s_{m}\otimes c_{m}),
\end{equation}
where $ \otimes$ denotes element-wise multiplication with broadcast.
Finally, the output will be added to get the fused feature $g^{AMFM}=g_{r}+g_{m}$.

\subsection{Temporal information enhanced module}
Although the appearance-motion fusion module can effectively fuse appearance information from the RGB image and motion information from the flow, we still observe undesirable prediction when we use it alone. We argue that this arises due to two reasons: 1)
the optical flow map can only provide temporal information between two adjacent frames, where no long-term temporal information exists;
2) as the flow feature is fused with the appearance feature in AMFM, some lower quality of flow may introduce extra noise to the network, leading to deteriorated predictions.

In order to solve this problem, we model long-term temporal information in our \enquote{Temporal information enhanced module} (TIEM) by adopting the bidirectional ConvLSTM \cite{song2018pyramid} to further constrain the cross-frames spatial and temporal information. Unlike previous methods \cite{fan2019shifting, yan2019semi}, which only add the temporal model in the highest level, we add a TIEM after each AMFM to promote information flow in each feature level between frames.

With the bidirectional ConvLSTM \cite{song2018pyramid}, we obtain hidden states $H^f_t$ and $H^b_t$ from both the forward and backward ConvLSTM units, which can be formulated as:
\begin{equation} 
H^{f}_{t} = ConvLSTM(H^{f}_{t-1}, g_{t}^{AMFM}; \gamma),
\end{equation}
\begin{equation} 
H^{b}_{t} = ConvLSTM(H^{b}_{t+1}, H^{f}_{t};\gamma),
\end{equation}
\begin{equation} 
s_{t}^{TIEM} = Conv(Cat(H^{f}_{t}, H^{b}_{t});\gamma),
\end{equation}
where $g_{t}^{AMFM}$ and $s_{t}^{TIEM}$ represent the features from the AMFM and TIEM respectively.

\subsection{Foreground-background similarity loss}

Different from \cite{zhang2020weakly} which can learn saliency from independent static images, our model learns video saliency with fixation-guided scribbles, where annotation of adjacent frames are related.
The large redundancy in adjacent frames makes it
possible to re-use scribble annotations of other frames to supervise the current frame. Further, we observe that it is difficult for the network to determine the category of each pixel without per-pixel annotation.
Motivated by \cite{yu2020context}, we propose our \enquote{Foreground-background similarity loss} to take advantage of limited weakly annotated labels and model the relationship of all points in adjacent frames. 
\begin{figure}[!t]
  \graphicspath{{figure1/network/}}
  \centering
  \includegraphics[width=0.70\linewidth]{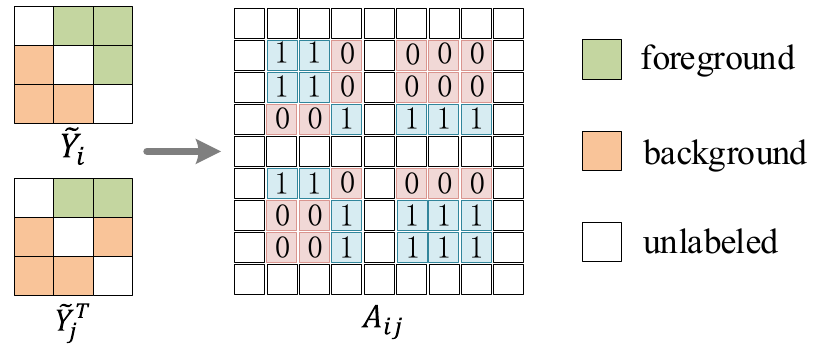}

  \caption{The illustration of how to get the ground truth of the similarity map $\hat{A}_{i,j}$. Note that, we do not define the similarity between unlabeled points and other points in $A_{i,j} $.}
  \label{gt}
\end{figure}
We argue that the similarity of features of the same category (both salient or both background) should be larger than that of different categories.
Based on this, we first calculate the similarity of two feature maps. To be specific, for the feature map of the $i$ th frame $f_i$ and $j$ th frame $f_j$, we first use a $1 \times 1$ convolutional layer to map them into an embedding space. Subsequently, we reshape them into $C \times WH$. Then we conduct matrix multiplication followed the sigmoid activation function $\sigma$ to get the similarity map $\hat{A}$ of $HW \times HW$ size. It can be formulated as:

\begin{equation} 
  \hat{A}_{i,j} = \sigma (Conv(f_i)^{T}Conv(f_j))
\end{equation}
\noindent
where $Conv(*)$ is a $1 \times 1$ convolutional layer. $\hat{A}_{i,j}$ represents the obtained similarity map between $i$ th frame and $j$ th frame.
Then we need to construct a ground truth map to supervise $\hat{A}_{i,j}$. Given the weakly annotated label of $i$ th frame $Y_i$, we first downsample it into the same size as the feature map $f_i$, so we obtain a smaller label $\tilde{Y_i}$. We encode the foreground part in $\tilde{Y_i}$ into [1, 0] and the background part into [0, 1], leading to a tensor $\tilde{Y_i}$ of size $2 \times H \times W$. Then, we reshape it into  $2 \times HW$. We do the same operations to the $j$-th frame and obtain $\tilde{Y_j}$. Then, we conduct the matrix multiplication again and obtain $A_{i,j} = \tilde{Y_i} \tilde{Y_j^T}$ of size $HW \times HW$. We visualize this process in Fig.~\ref{gt}. Note that all operations above are done on labeled points, which means that we do not define the similarity between unlabeled points and other points. We use $J$ to represent the set of points in $A_{i,j}$.
Then we can adapt partial cross-entropy loss \cite{tang2018normalized} to supervise the similarity map:
\begin{equation} 
  L_s^{i,j} = -\sum_{u,v \in J}(A_{u,v} \log \hat{A}_{u,v} + (1-A_{u,v}) \log (1-\hat{A}_{u,v})).
\end{equation}

For each iteration we have T frames, we can calculate the similarity loss for the current frame with other frames and itself. So the total similarity loss can be formulated as:

\begin{equation} 
  L_s = \sum_{i=1}^{T}\sum_{j=i}^{T} L_s^{i,j}.
\end{equation}

\subsection{Loss Function}
As shown in Fig.~\ref{network_overview}, we employ both partial cross-entropy loss $L_c$ and the proposed foreground-background similarity loss
$L_s$ to train our model. Apart from this, the gated structure-aware loss $L_g$ and edge loss $l_e$ proposed in \cite{zhang2020weakly} are also used. Note that, we do not show $L_g$ and $l_e$ in Fig.~\ref{network_overview} for simplicity. Both $L_s$, $L_g$ and $l_e$ are the loss for learning from scribble labels of the static image and $L_s$ is the loss for learning from a series of frames. Following the conventional video saliency detection pipeline, we pretrain with an RGB saliency dataset. Differently, we use the scribble annotation based dataset, namely S-DUTS \cite{zhang2020weakly}. Then, we finetine the network with our fixation guided scribble annotation.
To pretrain the network, we define the loss as:
\begin{equation}
    L_{pretrain} = \beta_{1} \cdot {L}_{c} + \beta_{3} \cdot {L}_{g} + \beta_{4} \cdot {L}_{e}.
\end{equation}
Then we finetune the network with loss function:
\begin{equation}
    L_{fine} = \beta_{1} \cdot {L}_{c} + \beta_{2} \cdot {L}_{s} + \beta_{3} \cdot {L}_{g} + \beta_{4} \cdot {L}_{e}.
\end{equation}
Empirically, we set $\beta_{1}=\beta_{2}=\beta_{4}=1$ and $\beta_{3}=0.3$.


\subsection{Saliency boosting strategy}
Our model based on the fixation guided scribble annotation leads to
competitive performance
as shown in Table \ref{tab:deep_unsuper_Performance_Comparison} \enquote{Ours}. Furthermore, we
notice that some SOD methods, \eg \cite{zhao2019egnet}, can also achieve reasonable results on VSOD datasets. Inspired by \cite{tang2018weakly},
we propose a saliency consistency based pseudo label boosting technique guided by the SOD model to further refine our annotation.

Specifically,
we adopt EGNet \cite{zhao2019egnet} to generate the saliency maps for the RGB images and optical flow of our video saliency training dataset, which are defined as $p_{rgb}$ and $p_{m}$ respectively. Note that choosing other off-the-shelf SOD methods is also reasonable. 
As done in \cite{li2020plug}, we choose the intersection of $p_{rgb}$ and $p_{m}$ as the fused saliency map $p = p_{rgb} \odot p_{m}$, which captures the consistent salient regions of $p_{rgb}$ and $p_{m}$.
Our basic assumption is that $p$ contains all the foreground scribble, and covers no background scribble.
With this, we define the quality score of $p$ as:
\begin{equation} 
score = \frac{\left\| T(p) \odot s_{fore} \right\|_{0}} {\left\|  s_{fore} \right\|_{0}} \cdot (1- \frac{\left\| T(p) \odot s_{back} \right\|_{0}} {\left\|  s_{back} \right\|_{0}})
\end{equation}
where $T(*)$ binarizes the saliency map with threshold 0.5.
$\| * \|_{0}$ is the L0 norm. $s_{fore}$ and $s_{back}$ are the foreground and background scribble respectively as shown in Fig.~\ref{fig:fixation_guided_scribble} (d).

The first
part of the quality score
aims to evaluate the coverage of foreground scribble over $p$,
while the latter encourages no background scribble to overlap $p$. In this way, the higher quality score indicates a better saliency map of $p$. We then choose saliency maps with quality score larger than a pre-defined threshold, \eg $Tr=0.98$.
\begin{figure}[!htp]
     \begin{center}
     \begin{tabular}{{c@{ } c@{ } c@{ } c@{ }}}
     {\includegraphics[width=0.23\linewidth]{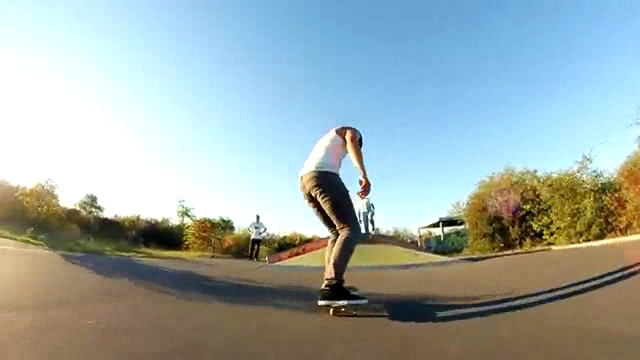}}&
      {\includegraphics[width=0.23\linewidth]{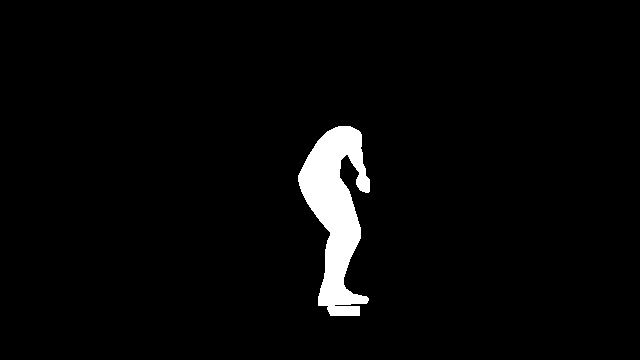}}&
      {\includegraphics[width=0.23\linewidth]{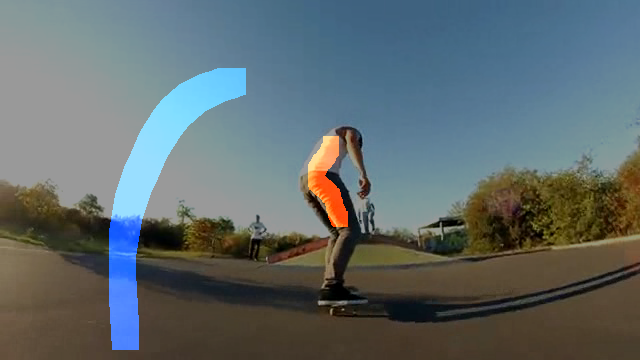}}&
      {\includegraphics[width=0.23\linewidth]{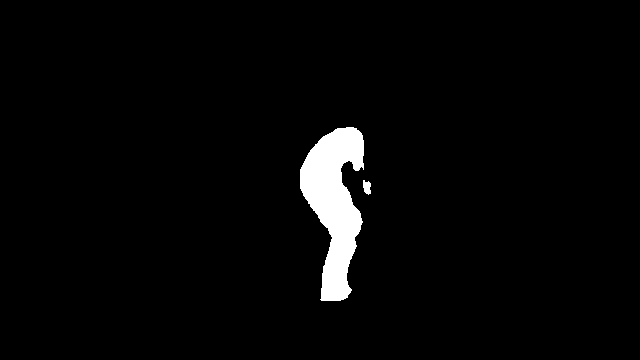}} \\
      \footnotesize{Image} & \footnotesize{GT} & \footnotesize{Scribble} & \footnotesize{Boosted} \\

      

     \end{tabular}
     \end{center}
     \caption{Our boosting strategy can refine the partial annotation.}
  \label{fig:boosting}
  \vspace{-4mm}
  \end{figure}
For each sequence, we can then obtain a set of high-quality pseudo saliency maps $P$. If the number of pseudo saliency maps in $P$ is larger than  10\% of the number of frames in current sequence, we replace the scribble annotation with the generated high quality pseudo label. Otherwise, we keep our original fixation guided scribble annotation for the entire sequence. We define the new weak annotation set as our first stage pseudo label set $D_{b1}$.

For sequences with high quality pseudo labels, we train them individually with the corresponding annotation in $D_{b1}$. After $K$\footnote{$K=8\times$ the size of the current sequence.} iterations of training, we perform inference using the trained model to obtain the second stage pseudo label set $D_{b2}$. Note that the model to train each sequence is introduced in Section \textcolor{red}{4.2} as the ablation study \enquote{B}.
Then, we treat $D_{b2}$ as our boosted annotation, and train our whole model with $D_{b2}$. During training, if a frame has a generated pseudo label, we directly use it as supervision. Otherwise, we use our scribble annotation to supervise. In Fig.~\ref{fig:boosting} we show the boosted annotation \enquote{Boosted}, which clearly show the effectiveness of our boosting strategy.



\section{Experimental Results}
\label{experimental_results}



\noindent\textbf{Dataset:}
Similar to the conventional VSOD learning pipeline,
our model is pre-trained on the scribble based image saliency dataset S-DUTS \cite{zhang2020weakly} and then fine-tuned on our fixation guided scribble annotations, namely the DAVIS-S and DAVSOD-S datasets. We evaluate the proposed method on six public datasets: VOS\cite{li2017benchmark}, DAVIS\cite{perazzi2016benchmark}, DAVSOD\cite{fan2019shifting}, FBMS\cite{ochs2013segmentation},  SegV2\cite{li2013video} and ViSal\cite{wang2015consistent}. The details of those datasets are shown in Table \ref{tab:dataset}.


\begin{table*}[t!]
  \centering
  \scriptsize
  \renewcommand{\arraystretch}{1.0}
  \renewcommand{\tabcolsep}{0.8mm}
 \caption{Benchmarking results. Bold numbers represent the best performance. $\uparrow \& \downarrow$ denote larger and smaller is better, respectively. Ours* means our method with the proposed boosting strategy. We use \red{red} and \blu{blue} to indicate the two best scores.}
  \begin{tabular}{lr|ccccccccccccc|ccc|ccc}
  \hline
  &  &\multicolumn{13}{c|}{Fully Sup. Models}&\multicolumn{5}{c}{Weakly/unsup. Models} \\
    & Metric
    & EGNet  & SCRN & PoolNet & SCOM & MBNM  & FCNS & PDB & FGRN & MGA & RCRNet &SSAV &PCSA
    &TENet  & SSOD  & GF  & SAG & Ours &Ours* \\ 
    & 
    & \cite{zhao2019egnet} &\cite{wu2019stacked} & \cite{liu2019simple}   & \cite{SCOM} & \cite{Li_2018_ECCV}& \cite{wang2017video} & \cite{song2018pyramid} & \cite{li2018flow} & \cite{li2019motion} &\cite{yan2019semi} &\cite{fan2019shifting} &\cite{gu2020pyramid}
    &\cite{ren2020tenet}  & \cite{zhang2020weakly}   & \cite{wang2015consistent}  & \cite{wang2015saliency} &   & \\ \hline
  \multirow{4}{*}{\textit{VOS}}
    & $S_{\alpha}\uparrow$    & 0.793 & 0.825 & 0.773 & 0.712 & 0.742 & 0.760 & 0.818  & 0.715 & 0.791 & \red{0.873} & 0.786  & 0.828   & \blu{0.845} & 0.682 & 0.615 & 0.619  & \blu{0.750} &  \red{0.765} 
    \\
    & $F_{\beta}\uparrow$    & 0.698 & 0.749 & 0.709 & 0.690 & 0.670 & 0.675  & 0.742 & 0.669  & 0.734 & \red{0.833} & 0.704  & 0.747    & \blu{0.781} & 0.648 & 0.506 & 0.482  & \blu{0.666}   & \red{0.702} \\
    & $\mathcal{M}\downarrow$ & 0.082 & 0.067 & 0.082 & 0.162 & 0.099 & 0.099 & 0.078 & 0.097  & 0.075 & \red{0.051} & 0.091  & 0.065     & \blu{0.052}  & 0.106 & 0.162 & 0.172  & \blu{0.091} & \red{0.089} \\ \hline
  \multirow{4}{*}{\textit{DAVIS}}
    & $S_{\alpha}\uparrow$    & 0.829 & 0.879 & 0.854    & 0.832 & 0.887 & 0.794 & 0.882 & 0.838   & \red{0.910}  & 0.886 & 0.892  & 0.902 & \blu{0.905}  & 0.795  & 0.688 & 0.676  & \blu{0.828}  &  \red{0.846}  \\
    & $F_{\beta}\uparrow$     & 0.768 & 0.847 & 0.815    & 0.783 & 0.861  & 0.708 & 0.855 & 0.783  & \red{0.892}  & 0.848 & 0.860  & 0.880 & \blu{0.881}  & 0.734  & 0.569 & 0.515 & \blu{0.779}   & \red{0.793}  \\
    & $\mathcal{M}\downarrow$ & 0.057 & 0.029 & 0.038   & 0.048 & 0.031 & 0.061 & 0.028 & 0.043    & 0.023  & 0.027 & 0.028  & \blu{0.022} & \red{0.017}  & 0.044  & 0.100   & 0.103  & \red{0.037} & \blu{0.038} \\ \hline
  \multirow{4}{*}{\textit{DAVSOD}}
    & $S_{\alpha}\uparrow$   & 0.719 & 0.745 & 0.702 & 0.599 & 0.637 & 0.657 & 0.698 & 0.693    & 0.741 & 0.741 & \blu{0.755}  & 0.741  & \red{0.779}    & 0.672 
    & 0.553  & 0.565  & \red{0.705} & \blu{0.694} \\
    & $F_{\beta}\uparrow$    & 0.604 & 0.652 & 0.592  & 0.464 & 0.520 & 0.521  & 0.572 & 0.573  & 0.643 & 0.654 & \blu{0.659}  & 0.656  & \red{0.697} & 0.556 
    & 0.334  & 0.370 & \red{0.605}  & \blu{0.593} \\
    & $\mathcal{M}\downarrow$ & 0.101 & 0.085 & 0.089 & 0.220 & 0.159 & 0.129 & 0.116 & 0.098   & \blu{0.083} & 0.087 & 0.084  & 0.086  & \red{0.070} & \red{0.101}  
    & 0.167   & 0.184 & \blu{0.103}  & 0.115  \\ \hline
  \multirow{4}{*}{\textit{FBMS}}
    & $S_{\alpha}\uparrow$    & 0.878 & 0.876 & 0.839 & 0.794 & 0.857 & 0.794 & 0.851 & 0.809  & \blu{0.908}  & 0.872 & 0.879 & 0.868 & \red{0.916}   & 0.747  & 0.651 & 0.659  & \blu{0.778}   & \red{0.803} \\
    & $F_{\beta}\uparrow$   & 0.848 & 0.861 & 0.830 & 0.797 & 0.816 & 0.759  & 0.821  & 0.767  & \blu{0.903}  & 0.859 & 0.865 & 0.837 & \red{0.915} & 0.727  & 0.571 & 0.564  & \blu{0.786}  & \red{0.792} \\
    & $\mathcal{M}\downarrow$  & 0.044 & 0.039 & 0.060 & 0.079 & 0.047 & 0.091 & 0.064 & 0.088 & \blu{0.027}  & 0.053 & 0.040 & 0.040 & \red{0.024}   & 0.083  & 0.160 & 0.161  & \red{0.072}  & \blu{0.073} 
  \\ \hline
   \multirow{4}{*}{\textit{SegV2}}
    & $S_{\alpha}\uparrow$    & 0.845 & 0.817 & 0.782 & 0.815 & 0.809 & * & 0.864 & * & \red{0.880} & 0.843 & 0.849 & 0.866     & \blu{0.868}       & 0.733  & 0.699 & 0.719  & \blu{0.804}  & \red{0.819} 
 \\
    & $F_{\beta}\uparrow$     & 0.774 & 0.760 & 0.704  & 0.764 & 0.716 & *  & 0.808 & *  & \red{0.829} & 0.782 & 0.797  & \blu{0.811} & 0.810       & 0.664 & 0.592 & 0.634  & \blu{0.738}  & \red{0.762} 
\\
    & $\mathcal{M}\downarrow$ & 0.024 & 0.025 & 0.025 & 0.030 & 0.026 & * & \blu{0.024} & * & 0.027 & 0.035 & \red{0.023} & \blu{0.024}     & 0.025       & \blu{0.039}  & 0.091 & 0.081  & \red{0.033}  & \red{0.033} 
 \\ \hline
   \multirow{4}{*}{\textit{ViSal}}
    & $S_{\alpha}\uparrow$   & 0.946 & 0.948 & 0.902 & 0.762 & 0.898 & 0.881 & 0.907& 0.861 & 0.940 & 0.922 & 0.942 & \blu{0.946} & \red{0.949}    & 0.853 & 0.757 & 0.749  & \blu{0.857}  & \red{0.883} 
 \\
    & $F_{\beta}\uparrow$    & 0.941 & 0.946 & 0.891  & 0.831 & 0.883 & 0.852  & 0.888  & 0.848  & 0.936 & 0.907 & 0.938 & \blu{0.941} & \red{0.949} & \blu{0.831}  & 0.683 & 0.688  & \blu{0.831}  & \red{0.875} 
\\
    & $\mathcal{M}\downarrow$ & 0.015 & 0.017 & 0.025 & 0.122 & 0.020 & 0.048 & 0.032 & 0.045 & \blu{0.017} & 0.026 & 0.021 & \blu{0.017} & \red{0.012}   & \blu{0.038}  & 0.107 & 0.105 & 0.041  & \red{0.035}
    
 \\
    \hline
  \end{tabular}
  \label{tab:deep_unsuper_Performance_Comparison}
\vspace{-2mm}
\end{table*}

\noindent\textbf{Implementation details:} As shown in Fig~\ref{network_overview}, our network takes image and flow as input. We first adopt an off-the-shelf optical ﬂow estimation method \cite{sun2018pwc} to compute the flow map from the previous frame to the current frame.
For the S-DUTS dataset, we just take the output from \cite{sun2018pwc} as the flow map of the static image by inputting the two same images into it. 
We take ResNet-50 \cite{he2016deep} pretrained on ImageNet \cite{deng2009imagenet} as the backbone. Note that during pretraining, there is no TIEM in our network. We resize all the frames to same spatial size of $256\times256$ before we feed them to the network. Random flipping and cropping are also added to avoid over-fitting.
The optimization algorithm is Adam
\cite{kingma2014adam} 
and the learning rate is 1e-4. We pre-train and fine-tune for 30 epochs and 20 epochs respectively. The batch size is set to one, and the length of frames per batch is set to four due to computation resource limitations. The whole pre-training and fine-tuning takes about four hours and nine hours respectively on a PC with an NVIDIA GeForce RTX 2080Ti GPU.
During test, our average processing time for one frame of a sequence is
0.035s.


\noindent\textbf{Competing methods:}
We compare our method with 16 state-of-the-art
image/video saliency methods as shown in Table~\ref{tab:deep_unsuper_Performance_Comparison}. Since SSOD \cite{zhang2020weakly} is the only scribble based saliency model,
we
finetune it with our scribble DAVSOD dataset for a fair comparison. 


\noindent\textbf{Evaluation metrics:}
We use three criteria to evaluate the performance of our method and competing methods, including Mean Absolute Error (MAE),  F-measure \cite{achanta2009frequency} ($F_{\beta}$), and the structure measure S-measure \cite{fan2017structure} ($S_\alpha$).





\subsection{Comparison with the state-of-the-art}
\noindent\textbf{Quantitative Comparison:}
In Table~\ref{tab:deep_unsuper_Performance_Comparison}, we show the results of our method and the competing methods. We can observe that our method can outperform all other weakly supervised or unsupervised method on six datasets. Comparing with the only scribble base method SSOD \cite{zhang2020weakly}, although it has been finetuned on DAVIS-S and DAVOSD-S, we still can surpasses it by a large margin. That is mainly because our method can take advantage of the motion and temporal information between frames. Moreover, our method can also achieve competitive performance with some fully-supervised methods, \eg, FCNS \cite{wang2017video}, FGRN \cite{li2018flow}, SCOM \cite{SCOM} and MBNM \cite{Li_2018_ECCV}.

\begin{figure*}[!htp]
   \begin{center}
   \begin{tabular}{{c@{ } c@{ } c@{ } c@{ } c@{ } c@{ } c@{ } c@{ } c@{ } c@{ }}}
    {\includegraphics[width=0.091\linewidth]{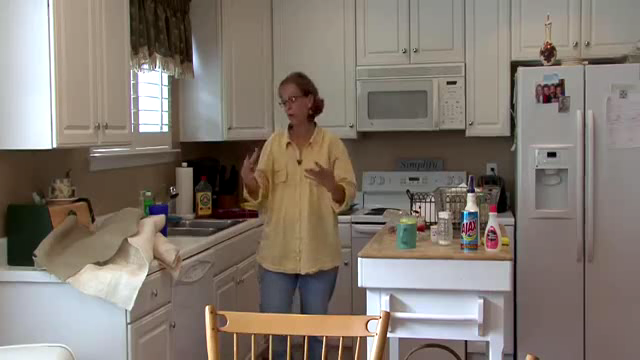}}&
    {\includegraphics[width=0.091\linewidth]{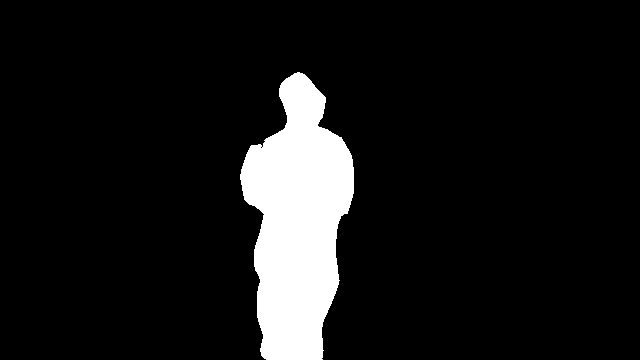}}&
    {\includegraphics[width=0.091\linewidth]{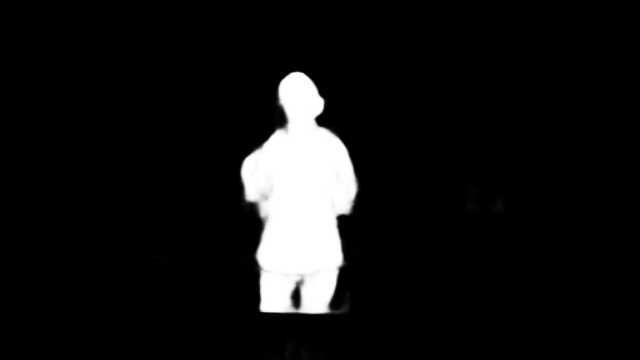}}&
    {\includegraphics[width=0.091\linewidth]{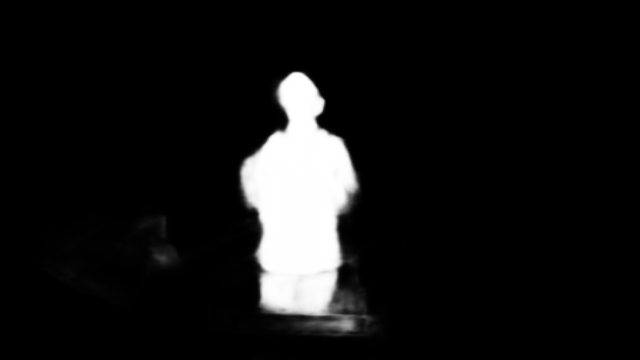}}&
    {\includegraphics[width=0.091\linewidth]{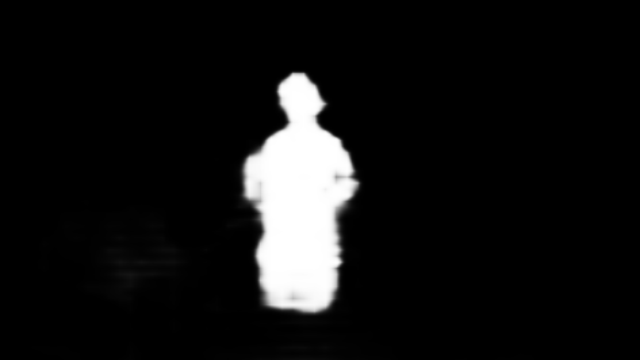}}&
    {\includegraphics[width=0.091\linewidth]{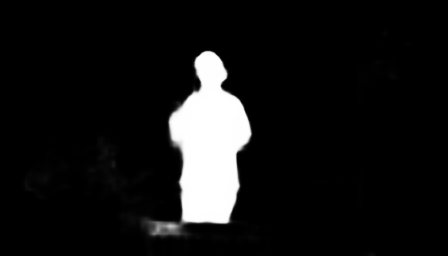}}&
    {\includegraphics[width=0.091\linewidth]{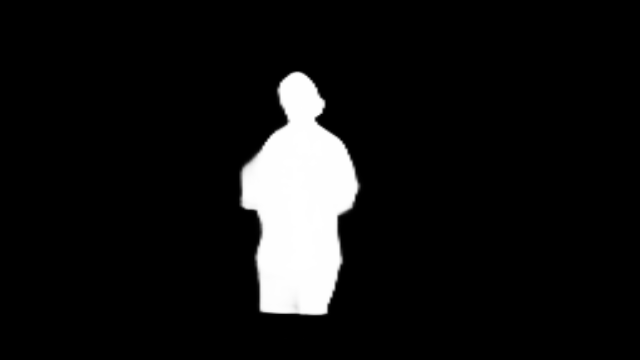}}&
    {\includegraphics[width=0.091\linewidth]{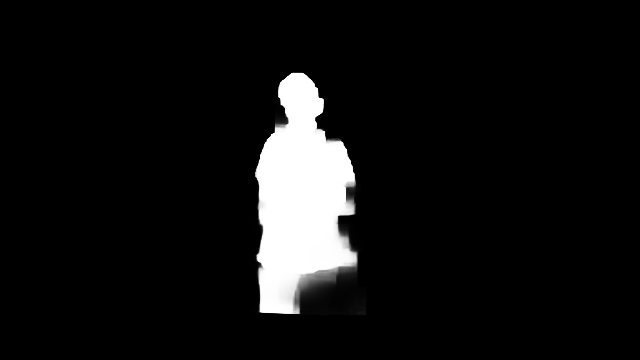}}&
    {\includegraphics[width=0.091\linewidth]{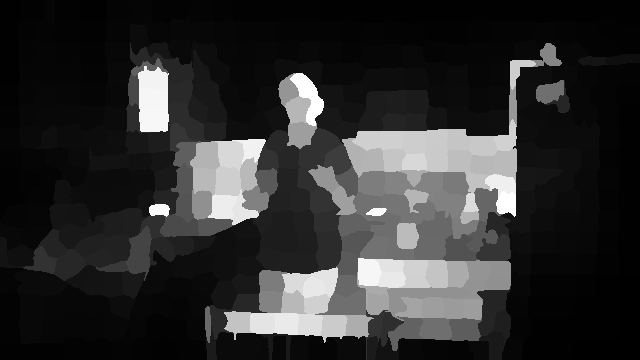}}&
    {\includegraphics[width=0.091\linewidth]{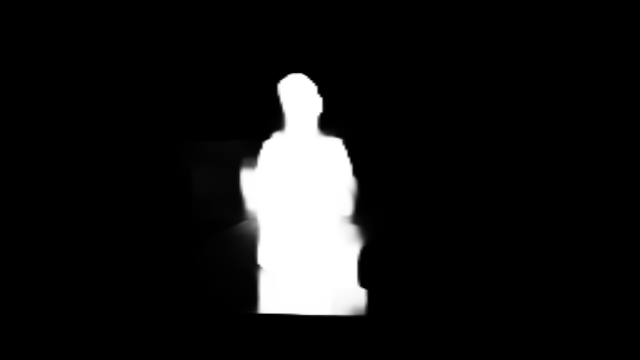}}
 \\

    {\includegraphics[width=0.091\linewidth]{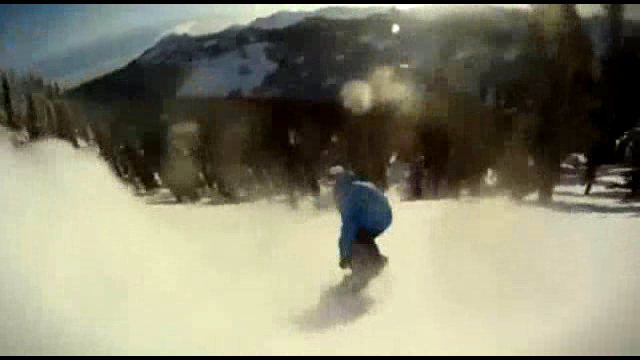}}&
    {\includegraphics[width=0.091\linewidth]{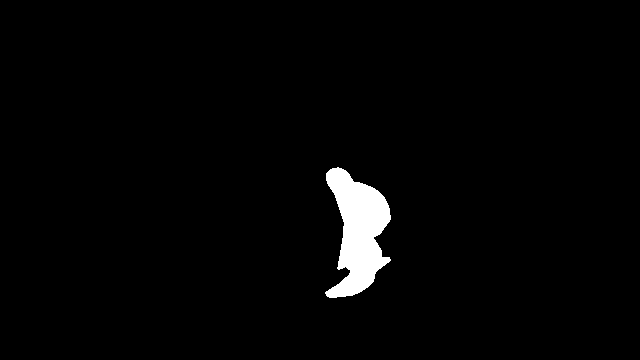}}&
    {\includegraphics[width=0.091\linewidth]{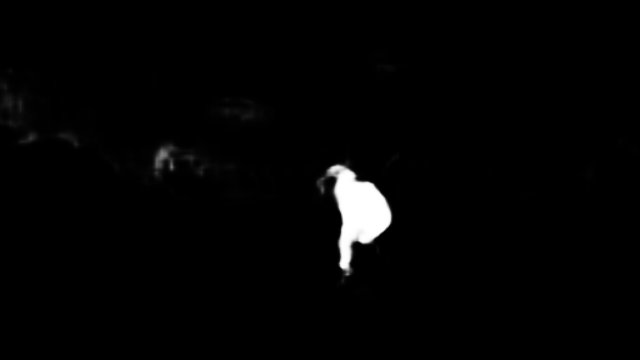}}&
    {\includegraphics[width=0.091\linewidth]{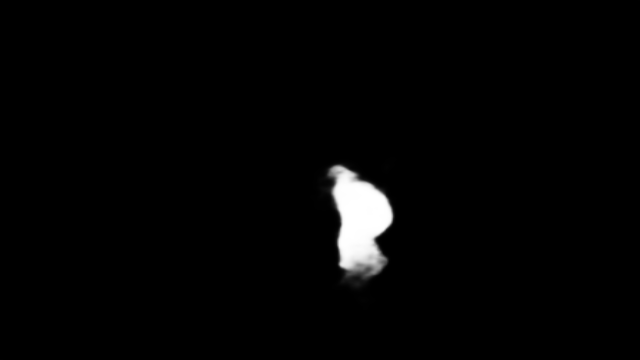}}&
    {\includegraphics[width=0.091\linewidth]{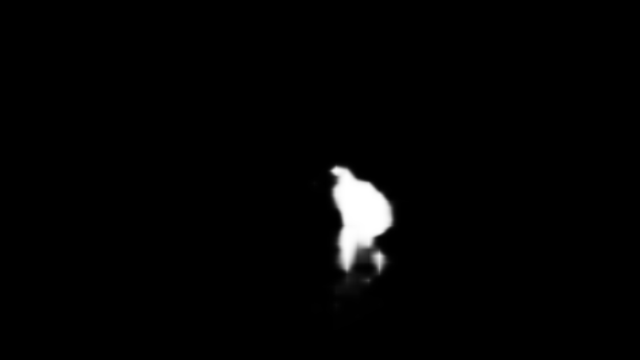}}&
    {\includegraphics[width=0.091\linewidth]{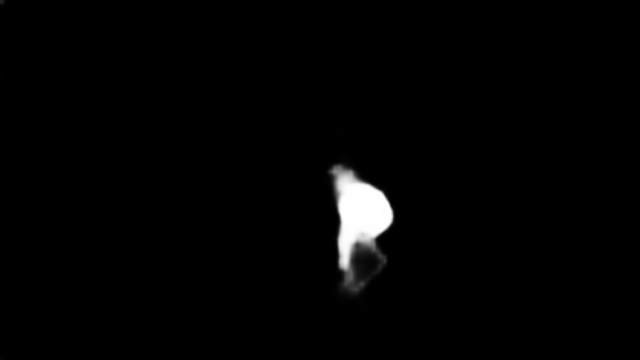}}&
    {\includegraphics[width=0.091\linewidth]{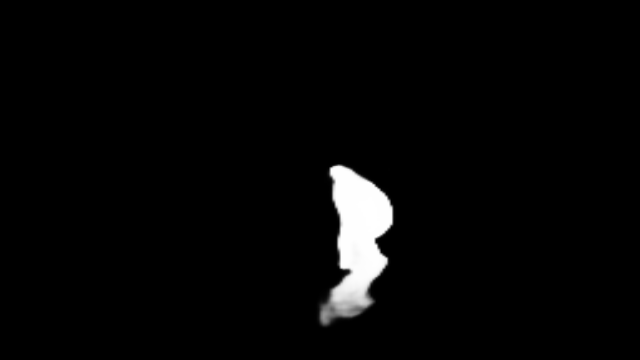}}&
    {\includegraphics[width=0.091\linewidth]{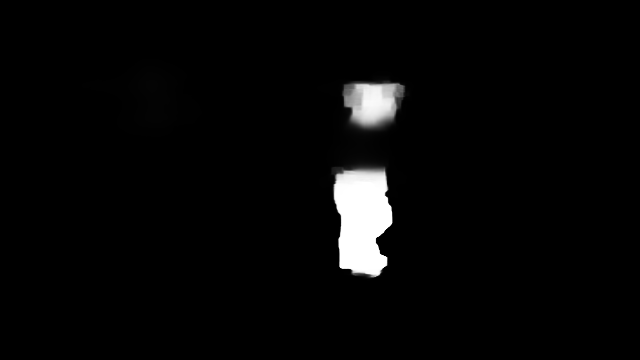}}&
    {\includegraphics[width=0.091\linewidth]{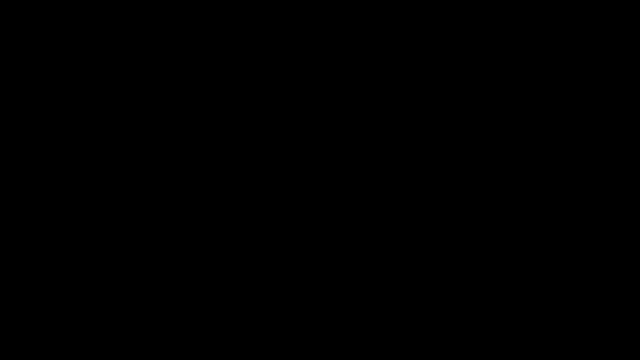}}&
    {\includegraphics[width=0.091\linewidth]{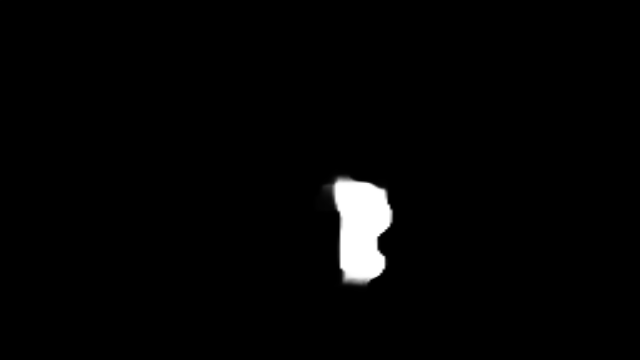}}
    \\
    {\includegraphics[width=0.091\linewidth]{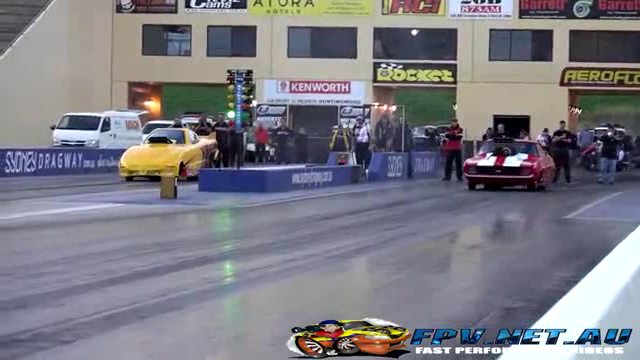}}&
    {\includegraphics[width=0.091\linewidth]{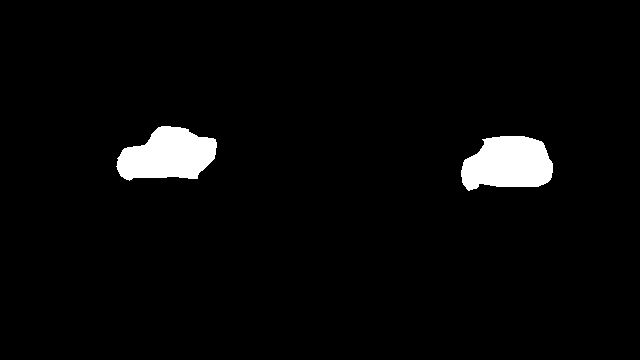}}&
    {\includegraphics[width=0.091\linewidth]{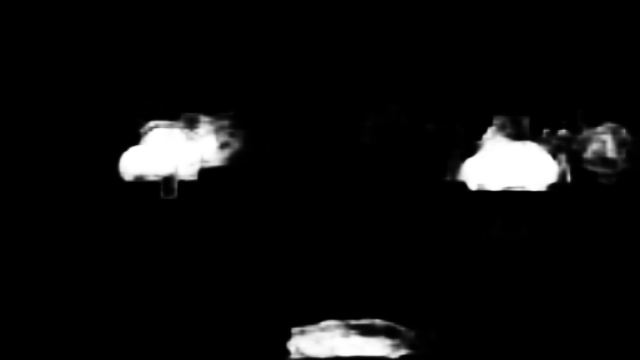}}&
    {\includegraphics[width=0.091\linewidth]{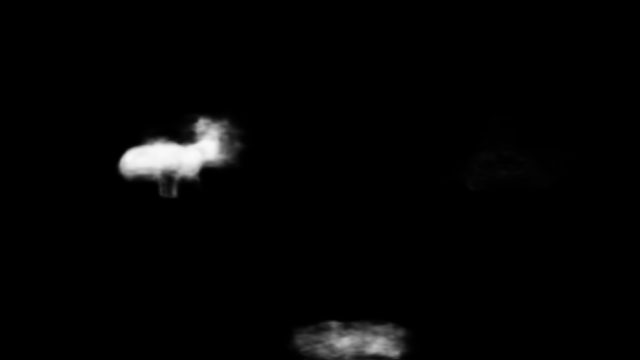}}&
    {\includegraphics[width=0.091\linewidth]{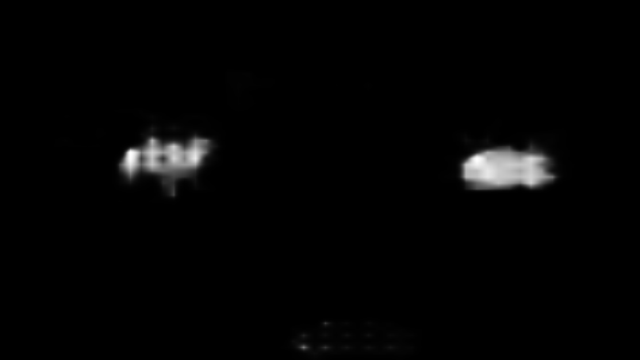}}&
    {\includegraphics[width=0.091\linewidth]{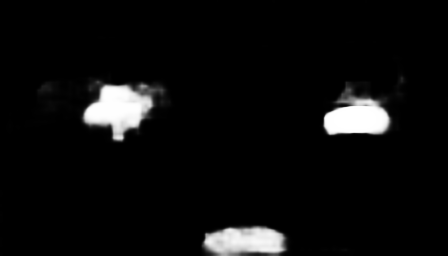}}&
    {\includegraphics[width=0.091\linewidth]{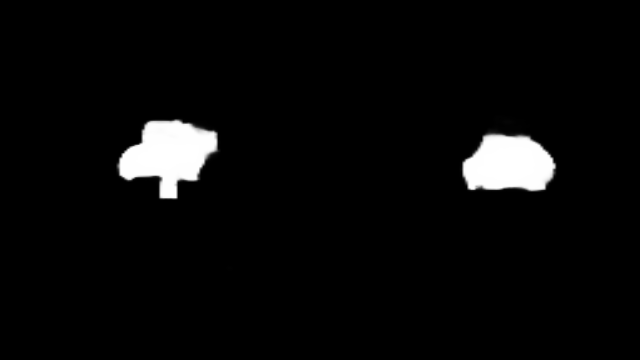}}&
    {\includegraphics[width=0.091\linewidth]{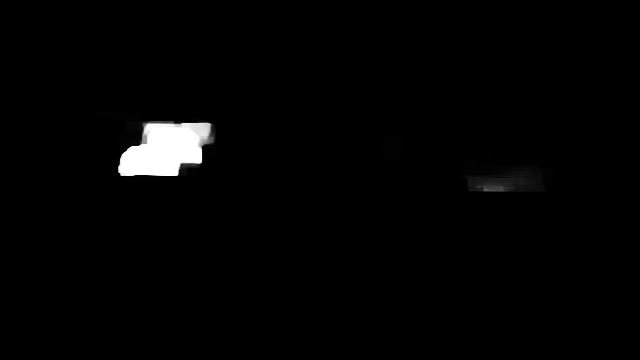}}&
    {\includegraphics[width=0.091\linewidth]{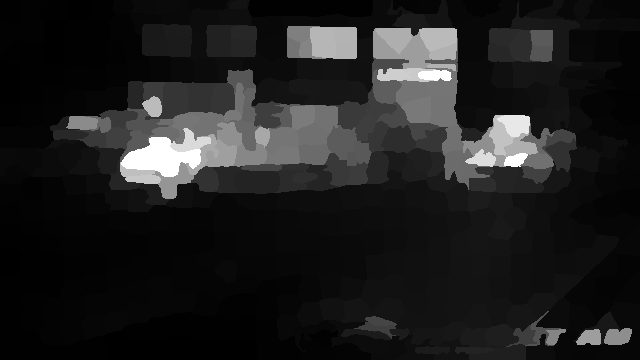}}&
    {\includegraphics[width=0.091\linewidth]{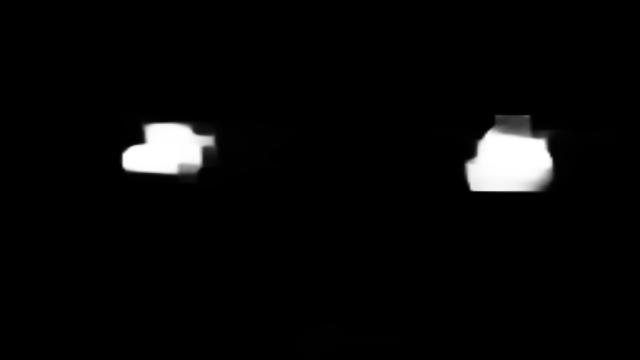}}
    \\
    
    {\includegraphics[width=0.091\linewidth]{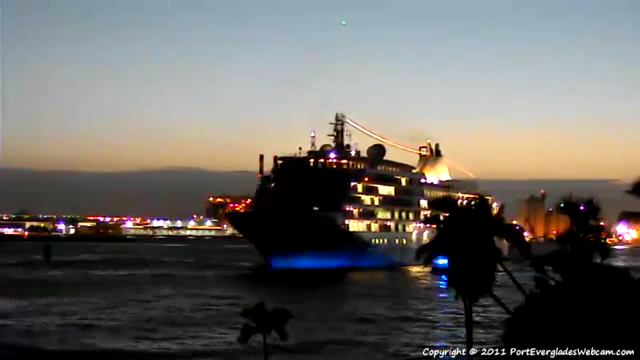}}&
    {\includegraphics[width=0.091\linewidth]{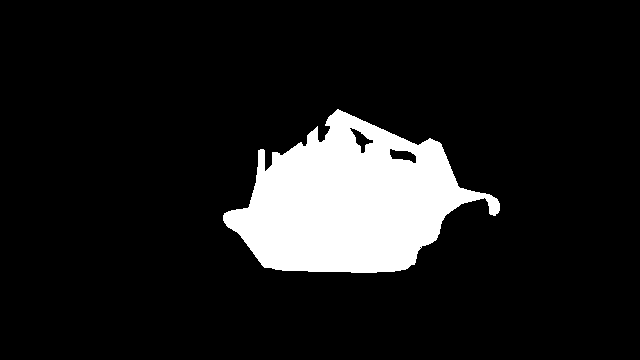}}&
    {\includegraphics[width=0.091\linewidth]{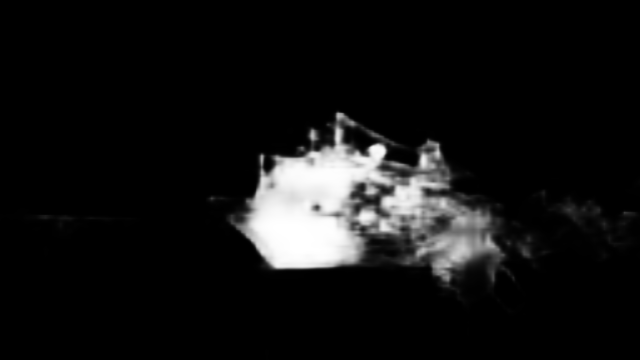}}&
    {\includegraphics[width=0.091\linewidth]{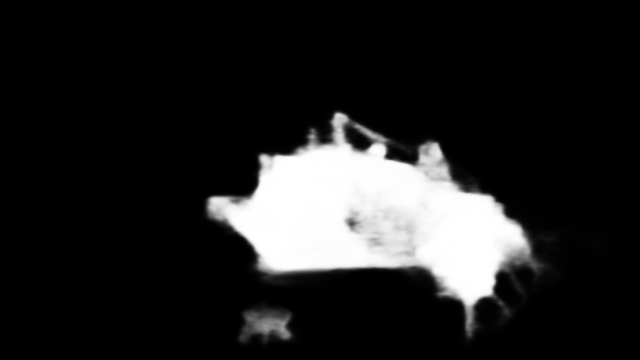}}&
    {\includegraphics[width=0.091\linewidth]{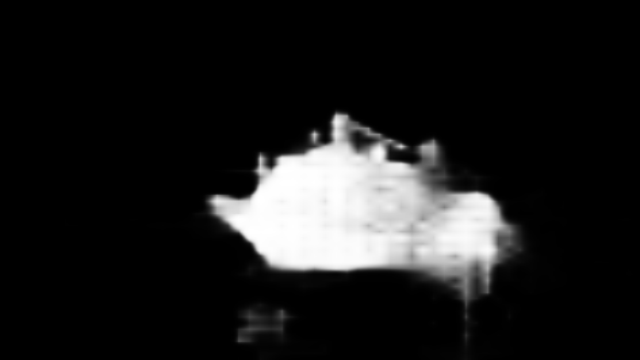}}&
    {\includegraphics[width=0.091\linewidth]{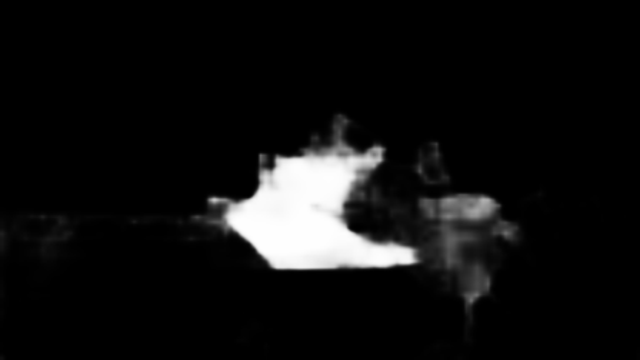}}&
    {\includegraphics[width=0.091\linewidth]{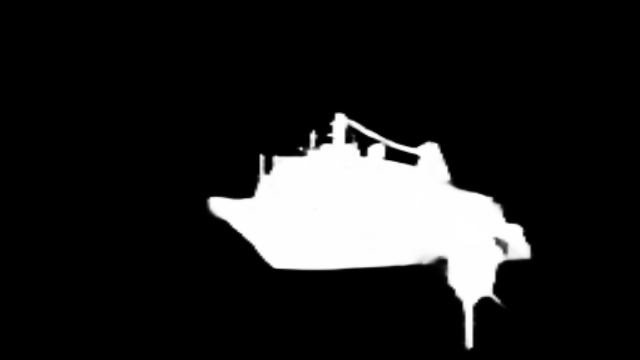}}&
    {\includegraphics[width=0.091\linewidth]{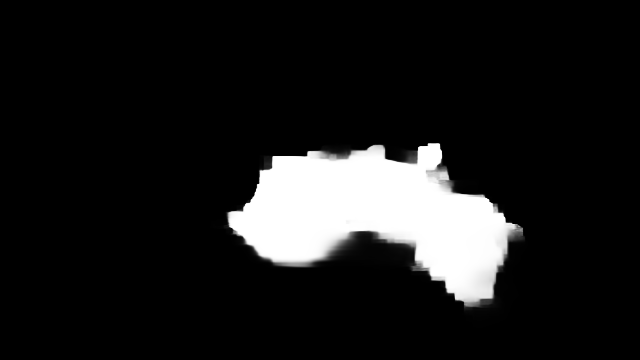}}&
    {\includegraphics[width=0.091\linewidth]{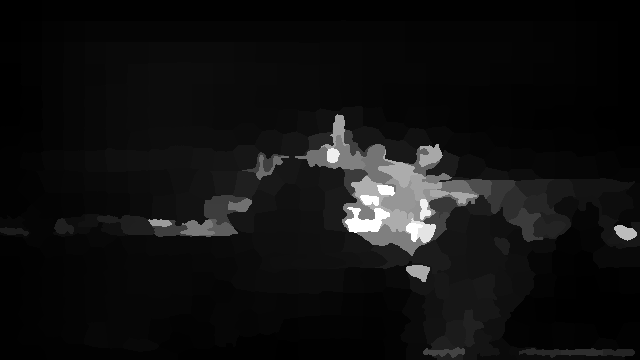}}&
    {\includegraphics[width=0.091\linewidth]{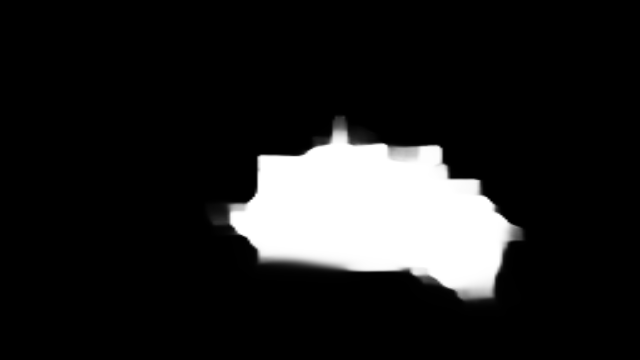}}
    \\

    \footnotesize{(a) Image} & \footnotesize{(b) GT} & \footnotesize{(c) MGA} & \footnotesize{(d) RCRNet} & \footnotesize{(e) SSAV} & \footnotesize{(f) PSCA} & \footnotesize{(g) TENet} & \footnotesize{(h) SSOD} & \footnotesize{(i) GF} & \footnotesize{(j) Ours} \\
   \end{tabular}
   \end{center}
   \caption{ Qualitative comparison with state-of-the-art video salient object detection methods.
}

\label{fig:qualitative_comparison}
\vspace{-3mm}
\end{figure*}

\noindent\textbf{Qualitative Comparison:}
We select four representative frames from four sequences in the testset of DAVSOD in Fig.~\ref{fig:qualitative_comparison}. We compare our method with the five best fully-supervised methods MGA \cite{li2019motion}, RCRNet \cite{yan2019semi}, SSAV \cite{fan2019shifting}, PSCA \cite{gu2020pyramid}, and TENet \cite{ren2020tenet} and two weakly/unsupervised methods SSOD \cite{zhang2020weakly} and GF \cite{wang2015consistent}. More qualitative comparison can be found in the supplementary materials. Benefiting from motion and temporal information, our method can locate salient objects more accurately than other weakly/unsupervised methods. Our method also shows comparable performance with fully-supervised methods on sequences with complex scenes (row 1), quick motion (row 2), multiple objects (row 3) and slow motion (row4).

\begin{table}[]
\centering
\scriptsize
\caption{Performance of our ablation study related experiments. 
}
\begin{tabular}{l|l|lll|lll}
\hline
\multicolumn{2}{l|}{\multirow{2}{*}{Method}} & \multicolumn{3}{l|}{DAVSOD}                              & \multicolumn{3}{l}{FBMS}                                \\ \cline{3-8} 
\multicolumn{2}{l|}{}                        & \multicolumn{1}{l|}{$S_{\alpha}\uparrow$} & \multicolumn{1}{l|}{$F_{\beta}\uparrow$} & $\mathcal{M}\downarrow$ & \multicolumn{1}{l|}{$S_{\alpha}\uparrow$} & \multicolumn{1}{l|}{$F_{\beta}\uparrow$} & $\mathcal{M}\downarrow$\\ \hline
\multicolumn{2}{l|}{B}    & 0.670  & 0.543 & 0.116 & 0.749 & 0.707 & 0.085 \\ 
\multicolumn{2}{l|}{B(G)}   & 0.578 & 0.424 & 0.222 & 0.631 & 0.568 & 0.196  \\ \hline
\multicolumn{2}{l|}{B-Fc}   & 0.669 & 0.538 & 0.123 & 0.769 & 0.743 & 0.081 \\
\multicolumn{2}{l|}{B-Fa}   & 0.678 & 0.556 & 0.119 & 0.763 & 0.735 & 0.084 \\
\multicolumn{2}{l|}{B-Fo}   & 0.682 & 0.562 & 0.112 & 0.775 & 0.763 & 0.08  \\ \hline
\multicolumn{2}{l|}{B-Fo-Th} & 0.681 & 0.556 & 0.121 & 0.780  & 0.769 & 0.076 \\
\multicolumn{2}{l|}{B-Fo-Ta} & 0.694 & 0.585 & 0.108 & \red{0.781} & 0.781 & 0.074 \\ \hline
\multicolumn{2}{l|}{B-Fo-Ta-L}      & \red{0.705} & \red{0.605} & \red{0.103} & 0.778 & \red{0.786} & \red{0.072} \\ \hline
\end{tabular}
\label{ablation_study_perfom}
\vspace{-3mm}
\end{table}

\subsection{Ablation study}
We thoroughly analyze the proposed framework and provide extra experiments as shown in
Table \ref{ablation_study_perfom}.\\
\noindent\textbf{Scribble annotation based baseline:}
We employ the proposed DAVSOD-S and DAVIS-S to finetune the base model, which has been pretrained on S-DUTS. The base model is constructed by removing all TIEM and replacing all AMFM with convolutional layers. It only takes RGB images as input. The performance is marked as \enquote{B}.
We also conduct an experiment by leveraging GraphCut \cite{boykov2001interactive} to generate masks given scribble annotations and directly adopting them to train \enquote{B}. This model is denoted as \enquote{B(G)}. The result in Table.~\ref{ablation_study_perfom} showers inferior performance of \enquote{B(G)}.
The main reason is graph-based algorithms can not generate accurate masks from simple scribble annotations, which further explains superior performance of the proposed solution.


\noindent\textbf{Different appearance-motion fusion strategy:}
In order to add the motion information, we introduce the optical flow map into the base model  and add the AMFM module into \enquote{B}. This model is named \enquote{B-Fo}. To demonstrate the effectiveness of the proposed appearance-motion fusion strategy, we also compare it with two simple fusion strategies: element-wise addition and concatenation. They are denoted as \enquote{B-Fa} and \enquote{B-Fc} respectively. As shown in Tab.~\ref{ablation_study_perfom}, our method surpasses \enquote{B-Fa} by 0.4\% on $S_{\alpha}$ and 0.6\% on $F_{\beta}$ on DAVSOD. The improvement on FBMS is much larger, compared with \enquote{B-Fc} and \enquote{B-Fa} the $F_{\beta}$ is increased by 2\% and 2.8\%, respectively.

\noindent\textbf{Temporal information enhanced module:}
We add TIEM to \enquote{B-Fo} to explore the effectiveness of long-term temporal information. Specifically, we propose two different solutions. Firstly, we only use the temporal model at the highest level of the network (Block 4 in Fig.~\ref{network_overview}), which leads to \enquote{B-Fo-Th}. Secondly, we introduce TIEM to every level of our network as in Fig.~\ref{network_overview}, which is \enquote{B-Fo-Ta}.
Experiments show that \enquote{B-Fo-Ta} works better than \enquote{B-Fo-Th}.
Especially on DAVSOD, we observe significant performance improvement of \enquote{B-Fo-Ta}, which achieves $\mathcal{M}$ of 0.108, far better than \enquote{B-Fo-Th} with $\mathcal{M}$ of 0.121.

\noindent\textbf{Foreground-background similarity loss:}
We add the foreground-background similarity loss to  \enquote{B-Fo-Ta} to cooperate with binary cross-entropy loss. The performance is indicated as \enquote{B-Fo-T-L}. Compared with \enquote{B-Fo-Ta}, since the proposed loss can provide extra frame-wise supervision, we obtain better performance with less training ambiguity.

\noindent\textbf{Boosting strategy:}
We perform the boosting strategy to our method in Tab.~\ref{tab:deep_unsuper_Performance_Comparison} \enquote{Ours}, and achieve
\enquote{Ours*}.
We observe that this strategy can generally improve model performance, which clearly shows the effectiveness of the proposed boosting strategy. Further, we notice decreased performance of \enquote{Ours*} compared with \enquote{Ours} on DAVSOD dataset. We then analyse the generated pseudo labels from the boosting strategy, and find that there exist some low quality pseudo labels, which mainly come from the inconsistent performance of EGNet \cite{zhao2019egnet} on our training dataset. This inspires us to explore further on boosting strategy. Designing a strategy to avoid the accumulated error due to boosting by taking both labels before and after the boosting strategy into consideration is a potential solution.
\section{Zero-shot Video Object Segmentation}
Similar to video salient object detection, the zero-shot video object segmentation aims to segment the
primary object in a video sequence, which is usually the salient object.
To evaluate generalization ability of the proposed method, we test on the validation set of DAVIS, which is widely used for zero-shot video object segmentation evaluation. We compare our method with three fully-supervised methods (PDB \cite{song2018pyramid}, AGNN \cite{wang2019zero}, MATNet \cite{zhou2020motion}) and four weakly/unsupervised methods (MM \cite{pathak2017learning}, TSN \cite{croitoru2017unsupervised}, COSE \cite{tsai2016semantic}, MuG \cite{lu2020learning}) on mean Jaccard index $\mathcal{J}$ and mean contour accuracy $\mathcal{F}$.
As shown in Table~\ref{VOS}, our method outperforms other weakly/unsupervised video segmentation methods on $\mathcal{J}$, with slightly decreased $\mathcal{F}$ compared with Mug\cite{lu2020learning}.
Note that Mug \cite{lu2020learning} is trained with more than 1.5 million frames, which is 100 times larger than our training dataset.
Furthermore, its inference time is 0.6 seconds per frame, and ours is 0.011 seconds per frame.
Both the decreased amount of training data and the efficient inference time indicate that the proposed weakly-supervised strategy has the potential to be applied to video object segmentation.

\begin{table}[t!]
  \centering
  \scriptsize
  \renewcommand{\arraystretch}{1.1}
  \renewcommand{\tabcolsep}{1.0mm}
 \caption{Performance of video object segmentation on DAVIS.}
  \begin{tabular}{lr|ccc|ccccc}
  \hline
  &  &\multicolumn{3}{c|}{Fully Sup. Models}&\multicolumn{5}{c}{Weakly/Un sup. Models} \\
    & Metric
    
 & PDB   &AGNN  &  MATNet   & MM   &TSN & COSE & MuG & Ours\\ 
    & 
 &\cite{song2018pyramid} &\cite{wang2019zero}  & \cite{zhou2020motion}   
 & \cite{pathak2017learning} &\cite{croitoru2017unsupervised} & \cite{tsai2016semantic} &\cite{lu2020learning}\\ \hline

  \multirow{2}{*}{\textit{DAVIS}}
    & $\mathcal{J}\uparrow$   & 77.2 & 80.7 & 82.4 & 48.9 & 31.2 & 52.8 & 61.2 & 63.8 \\
    & $\mathcal{F}\uparrow$   & 74.5 & 79.1 & 80.7 & 39.1 & 32.2 & 49.3 & 56.1 & 52.4 \\

    \hline
  \end{tabular}
  \label{VOS}
  \vspace{-4mm}

\end{table}

\section{Conclusion}
We propose a novel weakly-supervised VSOD network trained on the proposed fixation guided scribble datasets, namely DAVIS-S and DAVSOD-S. We introduce multi-modality learning and a temporal constraint
to effectively model spatio-temporal information. Furthermore, we propose a novel similarity loss and fully explore the limited weakly annotations. A saliency boosting strategy is also introduced to leverage off-the-shelf SOD methods. Extensive experiments on VSOD and VOS illustrate that our method is effective and general. Moreover, we are the first to use a weakly-supervised setting and achieve comparable results, which we hope may be inspiring for future work.

\section*{Acknowledgements}

\footnotesize{
This research was supported in part by the National Science Foundation of China under Grant U1801265 and
CSIRO's Machine Learning and Artificial Intelligence Future Science Platform (MLAI FSP).}

{\small
\bibliographystyle{ieee_fullname}
\bibliography{egbib}
}
\end{document}